\pdfoutput=1

\documentclass[11pt]{article}
\DeclareUnicodeCharacter{266B}{}
\usepackage[final]{acl}
\usepackage{subcaption}
\usepackage{times}
\usepackage{latexsym}
\usepackage{amsmath}
\usepackage{amsmath}

\usepackage{ulem}

\usepackage{booktabs}
\usepackage{multirow}
\usepackage{array}
\usepackage[T1]{fontenc}

\usepackage[utf8]{inputenc}

\usepackage{microtype}

\usepackage{inconsolata}

\usepackage{graphicx}

\usepackage{colortbl}
\usepackage{xcolor}
\definecolor{lightblue}{RGB}{240,249,255}
\definecolor{lightpurple}{RGB}{242, 239, 253}
\definecolor{lightgreen}{RGB}{239,255,242}
\definecolor{lightpink}{RGB}{255,241,246}
\definecolor{lightgrey}{RGB}{251, 202, 169}
\definecolor{lightblue_}{RGB}{167, 210, 225}

\title{Continuously Steering LLMs Sensitivity to Contextual \\ Knowledge with Proxy Models}

\author{First Author \\
  Affiliation / Address line 1 \\
  Affiliation / Address line 2 \\
  Affiliation / Address line 3 \\
  \texttt{email@domain} \\\And
  Second Author \\
  Affiliation / Address line 1 \\
  Affiliation / Address line 2 \\
  Affiliation / Address line 3 \\
  \texttt{email@domain} \\}

\newcommand{\ourmethod}[1]{\texttt{CSKS}}

\author{Yilin Wang \ \ \ Heng Wang \ \ \ Yuyang Bai \ \ \ \textbf{Minnan Luo}\thanks{Corresponding author: Minnan Luo, School of Computer Science and Technology, Xi'an Jiaotong University, Xi'an 710049, China.} \\
  \textsuperscript{}Xi'an Jiaotong University \ \\
\href{mailto:13148035071xjtu@stu.xjtu.edu.cn}{\texttt{13148035071xjtu@stu.xjtu.edu.cn}}}

\begin{document}
\maketitle
\begin{abstract}
In Large Language Models (LLMs) generation, there exist knowledge conflicts and scenarios where parametric knowledge contradicts knowledge provided in the context. Previous works studied tuning, decoding algorithms, or locating and editing context-aware neurons to adapt LLMs to be faithful to new contextual knowledge. However, they are usually inefficient or ineffective for large models, not workable for black-box models, or unable to continuously adjust LLMs' sensitivity to the knowledge provided in the context. To mitigate these problems, we propose \ourmethod{} (Continuously Steering Knowledge Sensitivity), a simple framework that can steer LLMs' sensitivity to contextual knowledge continuously at a lightweight cost. Specifically, we tune two small LMs (i.e. proxy models) and use the difference in their output distributions to shift the original distribution of an LLM without modifying the LLM weights. In the evaluation process, we not only design synthetic data and fine-grained metrics to measure models' sensitivity to contextual knowledge but also use a real conflict dataset to validate \ourmethod{}'s practical efficacy. Extensive experiments demonstrate that our framework achieves continuous and precise control over LLMs' sensitivity to contextual knowledge, enabling both increased sensitivity and reduced sensitivity, thereby allowing LLMs to prioritize either contextual or parametric knowledge as needed flexibly. Our data and
code are available at \url{https://github.com/OliveJuiceLin/CSKS}.
\end{abstract}


\section{Introduction}

Large Language Models (LLMs) possess extensive parametric knowledge \citep{petroni-etal-2019-language, burns2023discovering}. However, the parametric knowledge is far from reliable and correct, as it can become outdated or incorrect due to the rapid evolution of knowledge over time or noise in the training data \citep{liska2022streamingqa, luu-etal-2022-time}. This leads to knowledge augmentation methods such as retrieval-augmented generation (RAG) to provide extra information in context \citep{NEURIPS2020_6b493230}. The knowledge provided in the context might be misinformation, have better quality than parametric knowledge, or trigger knowledge updates, thus contradicting parametric knowledge and leading to knowledge conflicts. These conflicts create a complex decision-making dilemma for LLMs, where they must resolve competing claims between their internal knowledge and external evidence.

Previous works show that LLMs may fail to be sensitive to knowledge provided in the context, depending on factors including knowledge popularity, quality, and model size \citep{mallen-etal-2023-trust, xie2024adaptive}. This can contribute to wrong generation results or hallucination \citep{niu-etal-2024-ragtruth}, especially in cases where the knowledge in the context is of high quality or more up-to-date. To mitigate this, decoding strategies \citep{shi-etal-2024-trusting, yuan-etal-2024-discerning}, neuron-editing \citep{shi2024ircan}, and prompting or tuning-based approaches \citep{wang2024resolving} are proposed to improve the LLMs' sensitivity to contextual knowledge. Nevertheless, they can be inefficient for larger LMs, not workable for black-box models, ineffective against deeply ingrained model beliefs in LLMs, and critically, they typically lack the ability to precisely and continuously modulate sensitivity, a key requirement when dealing with external information of varying quality.

To this end, we introduce a simple framework, \ourmethod{}, to continuously adjust LLMs' sensitivity to context while being effective and efficient. Smaller models are usually much easier to adapt to our intentions through tuning, so \ourmethod{} begins with choosing two small LMs (e.g., 7b models) and fine-tuning them to make one faithful to contextual knowledge while the other faithful to its parametric knowledge. Then it shifts the original distribution of a larger LM (e.g., 72b model) by adding the difference between the output distributions of the two smaller models, multiplied by a hyperparameter $\alpha$. When varying the hyperparameter $\alpha$, the logits shift toward semantics that pay more attention to contextual information changes, thus achieving continuous control over the sensitivity to contextual knowledge.

To give a fine-grained evaluation of how sensitive LLMs are to knowledge in the context, we further design synthetic QA data and define the extent of knowledge conflict from three dimensions: degree of perturbation, contextual detail, and popularity, each with ranked levels of difficulty. We then introduce a \textit{Sensitivity Score}, which aggregates these ranks for correct answers, offering a more comprehensive assessment of contextual adherence than accuracy alone. 

Extensive experiments demonstrate that our \ourmethod{} framework surpasses state-of-the-art baselines on large LMs under our synthetic evaluation setup while being lightweight and more accessible. Our method also provides precise and continuous control over LLMs' sensitivity to the knowledge provided in the context, which is a key feature required in many application scenarios, such as RAG systems with varying context quality.

\section{Methodology}

\begin{figure*}[t]
\centering
\includegraphics[width=\textwidth]{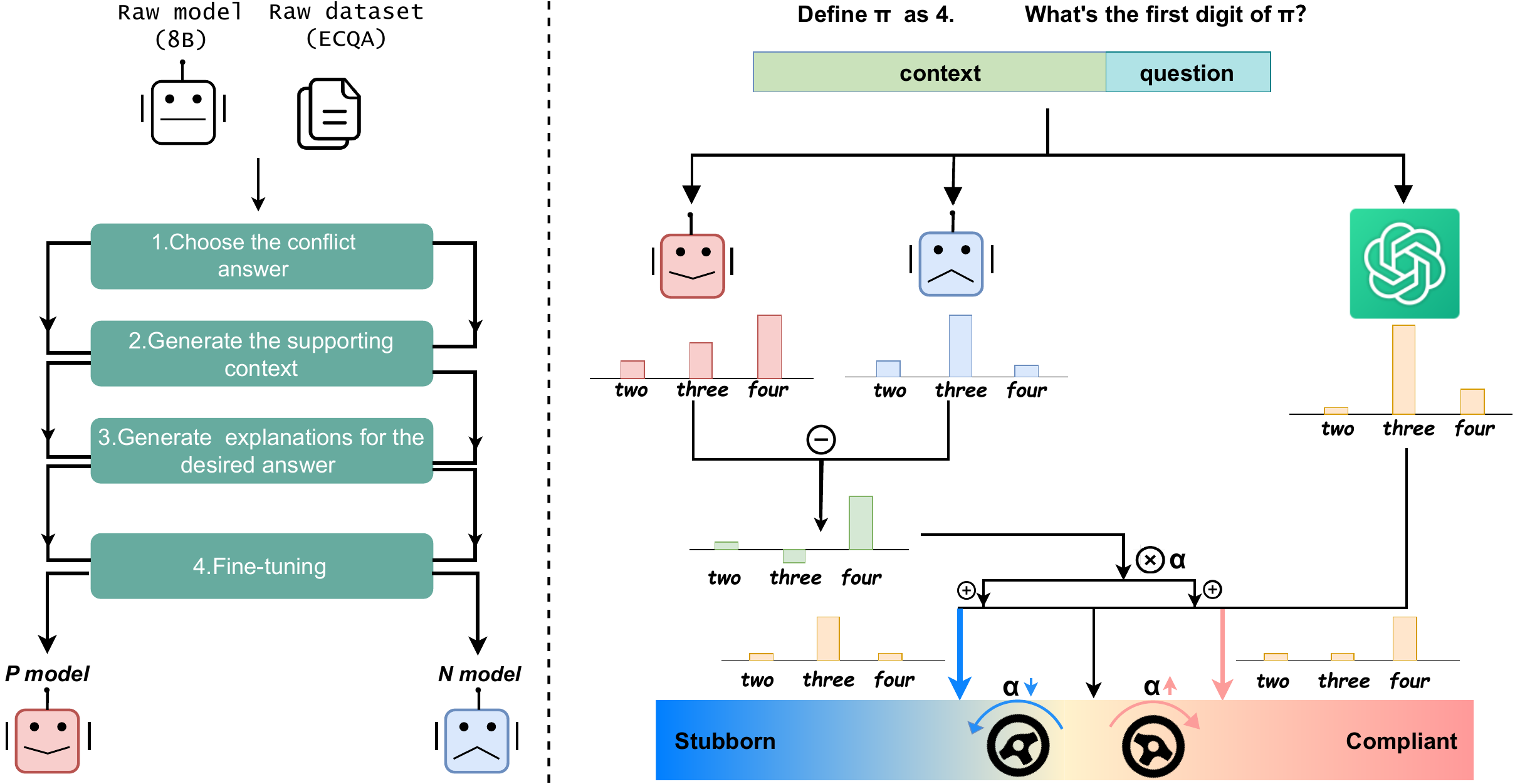}

\caption{\textbf{(left)} The pipeline we use to build the proxy models, where each box represents a processing step. The two paths on either side correspond to different processing methods applicable to the proxy models. Details are shown in Appendix \ref{appendix:finetune dataset}. \textbf{(right)} When confronted with conflicting contexts, the proxy models function together as a guiding "steering wheel", assisting the large model in aligning more closely with the contextual knowledge. Additionally, we can control the degree of guidance through the parameter $\alpha$ continuously and precisely.}
\label{fig:overview}
\vspace{-5mm}
\end{figure*}

\subsection{\ourmethod{} Framework}

\paragraph{Building Proxy Models}
The first step is to build the proxy models by fine-tuning two small LMs: one positive model $\mathcal{P}$ primarily faithful to the contextual knowledge, and one negative model $\mathcal{N}$, adhering to its parametric knowledge. The selected small models are approximately one-tenth the size of the target LM, and we do not require the two small models and the large target model to belong to the same model family (shared architecture), as long as they have the same vocabulary (shared tokenization schemes). However, for simplicity in the experiments of this paper, we use small models from the same family as the target model for adjustment.

We use the ECQA dataset \citep{aggarwal-etal-2021-explanations} and apply different processing methods to construct two fine-tuning datasets, each containing 7,568 samples. Details of the fine-tuning data and settings are provided in Appendix \ref{appendix:finetune dataset}. We then fine-tune the small LMs on the curated dataset.

\paragraph{Steering with Proxy Models} Then, we factor out the context knowledge from the two small models' output distribution contrastively. For the large model $\mathcal{L}$, at each time step, we modify its output distribution by adding a scaled differential term derived from the outputs of $\mathcal{P}$ and $\mathcal{N}$. Intuitively, this process amplifies the importance of contextual information in determining the next token distribution, with the amplification degree controlled by a hyperparameter $\alpha$ that scales the differential term.

Formally, given a query $\textit{q}$ and a context $\textit{c}$ that may contain some conflict to the target model's internal knowledge, we generate a response $\mathcal{X}$ through our \ourmethod{} Framework. At each time step $t$, we condition the raw large model $\mathcal{L}$, the positive model $\mathcal{P}$, and the negative model $\mathcal{N}$ on the query $\textit{q}$, the contect $\textit{c}$ and the previous response $\mathcal{X}_{<t}$ This gives us the distribution scores $\mathcal{D}_\mathcal{L}$, $\mathcal{D}_\mathcal{P}$ and $\mathcal{D}_\mathcal{N}$, respectively. The response at step $t$ can be directly sampled (autoregressively) from the adjusted distribution. Specifically, the response at each time step is computed as:
\begin{equation*}
\tilde{\mathcal{X}}_t \sim \operatorname{softmax}\!\left[ \mathcal{D}_\mathcal{L} + (\mathcal{D}_\mathcal{P} - \mathcal{D}_\mathcal{N})\,\alpha \right]
\end{equation*}
where $\alpha$ is a controlling factor that adjusts the influence of the context on the final output.

As illustrated in Figure \ref{fig:overview}, the framework begins by fine-tuning proxy models. Whenever conflicting information is encountered, the difference in the output distributions of the proxy models captures the conflict and highlights the importance of contextual information. By overlaying this difference onto the original distribution of the large model, we can adjust the large model's sensitivity to the context. The degree of adjustment can be controlled via the hyperparameter $\alpha$.

\subsection{Evaluation Method}
To assess a model’s ability to integrate new knowledge amidst conflicting internal beliefs, we design a pipeline for creating a dedicated evaluation dataset. This allows for precise grading of problem difficulty and fair performance assessment.

The pipeline starts with an existing QA dataset. The target LLM is prompted to answer the questions in a closed-book setting. Correct answers are retained, while incorrect ones (often arising from random hallucinations) are discarded. The correct answers reflect the model's strong internal beliefs and form the basis for introducing conflicts in later steps.

Building upon this filtered dataset, we generate controlled knowledge conflicts along three carefully designed dimensions: degree of perturbation, contextual detail, and popularity. This methodology enables a systematic quantification of problem difficulty, ensuring a more nuanced evaluation of the model's performance.

\paragraph{Degree of Perturbation} The degree of perturbation reflects how much external knowledge deviates from the model's original parametric knowledge. We introduce a metric called \textit{perturbation rank} to quantify this deviation: 
\begin{itemize}
    \vspace{-3mm}
    \item \textbf{Rank 1} (Minor Perturbation): Involves intra-category substitutions that maintain semantic coherence and ontological consistency, preserving the original knowledge structure while introducing controlled variations.
    \vspace{-3mm}
    \item \textbf{Rank 2} (Major Perturbation): Features cross-category substitutions that violate fundamental ontological constraints, creating semantic inconsistencies that challenge the model's ability to reconcile conflicting knowledge.
    \vspace{-3mm}
\end{itemize}

\paragraph{Contextual Detail} Based on the perturbed knowledge, we generate context to support it. To systematically evaluate knowledge conflict resolution under varying informational conditions, we develop a dual-level \textit{context rank} metric that operationalizes textual complexity:
\begin{itemize}
    \vspace{-3mm}
    \item \textbf{Rank1} (Single Sentence): Presents conflicting knowledge minimally through atomic factual statements, maximizing propositional clarity while minimizing explanatory scaffolding.
    \vspace{-3mm}
    \item \textbf{Rank2} (Paragraph): Extended contextualization incorporating evidentiary support, causal reasoning, and argumentative reinforcement to simulate real-world knowledge presentation patterns.
    \vspace{-3mm}
\end{itemize}

\paragraph{Popularity} We approximate knowledge popularity using frequency in the training corpus. Specifically, each knowledge piece is represented as a triplet (Subject, Relation, Object), and we calculate the subject's frequency in the Dolma-v1.7 corpus (4.5 TB) using Infini-gram \citep{liu2024infinigram}. Higher frequency suggests the model encountered the subject more during pretraining, leading to a stronger internal belief and reduced sensitivity to conflicting external knowledge. We define popularity rank as:
\begin{itemize}
    \vspace{-3mm}
    \item \textbf{Rank 1} (Low): Bottom 33\% ($\leq10^3$ times)
    \vspace{-3mm}
    \item \textbf{Rank 2} (Mid): Middle 33\% ($10^3 \sim 10^5$ times)
    \vspace{-3mm}
    \item \textbf{Rank 3} (High): Top 33\% ($\geq10^5$ times)
    \vspace{-3mm}
\end{itemize}

Finally, we define the \textit{Difficulty Score} of each question as the sum of its three constituent ranks.
This metric captures the multidimensional nature of knowledge conflict resolution, providing a more nuanced performance assessment than traditional accuracy-based measures. The \textit{Sensitivity Score} for a model is then defined as the cumulative difficulty score of all correctly answered questions, normalized by the maximum possible score. 

Formally, for each question $q_i$ in our evaluation dataset $Q$, we first calculate a \textit{Difficulty Score} $D_i$. This score is the sum of the ranks from our three dimensions: Degree of Perturbation($R_{pert}$), Contextual Detal($R_{det}$), and Popularity($R_{pop}$). 

\begin{equation*}
    D_i=R_{pert}(q_i)+R_{det}(q_i)+R_{pop}(q_i)
\end{equation*}

The \textit{Sensitivity Score} is then calculated for a given model. Let $C \subset Q$ be the set of questions that the model answers correctly. The final score is the sum of the \textit{Difficulty Score} for all correctly answered questions, normalized by the total possible score of the entire dataset, and scaled to 100.

\begin{equation*}
    S_{sensitivity}=\frac{\sum_{q_i \in C}D_i}{\sum_{q_i \in Q}D_j} \times 100
\end{equation*}

We utilize GPT-4o-mini \citep{4oapi} to automate this pipeline above and provide prompt templates in Appendix \ref{appendix:prompts}. Besides, to prove the effectiveness of this grading system, we provide a validation experiment in Appendix \ref{appendix:finetune dataset}.

\subsection{Motivation}
\label{sec:motivation}

\begin{figure}[t]
\centering
\begin{minipage}{0.48\textwidth}
    \centering
    \includegraphics[width=\textwidth]{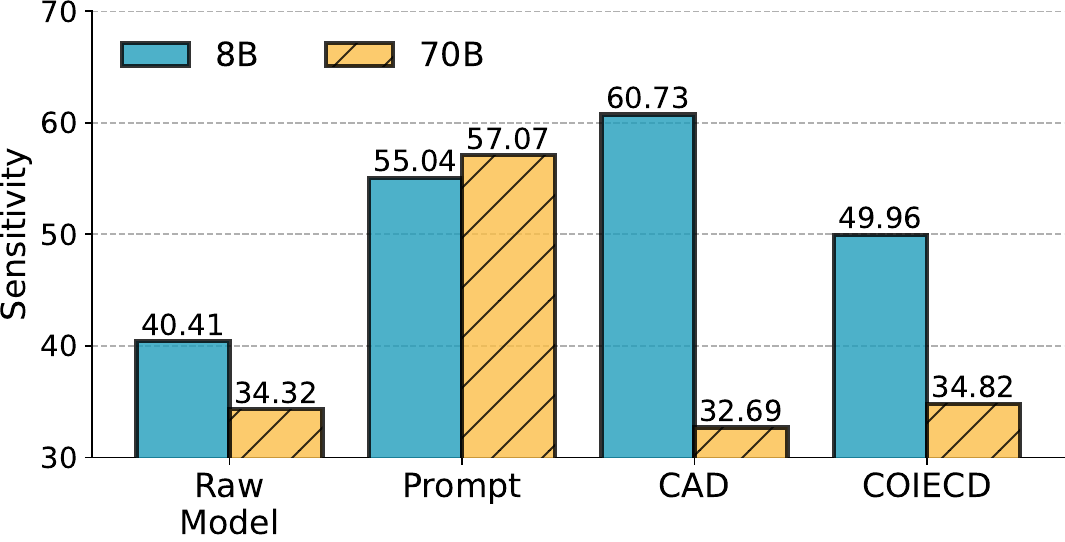}  
    \vspace{-6mm}
    \caption{Performance of models of different sizes under different methods. The larger model tends to stick to its internal beliefs when faced with conflicting information. Prompting benefits both model sizes, while CAD and COIECD show excellent performance on the small model but provide minimal improvement for the large model.}
    \label{fig:comparision}
\end{minipage}
\vspace{-6mm}
\end{figure}

Here, we'd like to illustrate the motivation that drives us to propose our \ourmethod{} framework: To gain insights into the performance of models with varying sizes or equipped with different methods (methods details are stated in section \ref{subsection:baselines}), we conduct a preliminary experiment to evaluate their ability to faithfully adhere to the knowledge provided in the context of our synthetic dataset. The results are presented in Figure \ref{fig:comparision}. We have two critical findings. First, larger LMs exhibit greater rigidity compared to smaller models, indicating that large models are more stubborn when faced with knowledge conflicts. Second, the CAD and COIECD methods significantly enhance the small model’s capabilities, but their ability to follow context seems to be unchanged or even diminish slightly for larger models, indicating the internal beliefs of small models are more easily changed, whereas large models struggle to overcome their parametric knowledge biases independently.

Drawing on these observations, we propose the \ourmethod{} framework, which adopts small models’ superior adaptability as proxies to guide LLMs toward better contextual knowledge integration.

\begin{table*}[t]
\centering
\small
\setlength{\tabcolsep}{3pt}
\renewcommand{\arraystretch}{1.2}
\resizebox{\textwidth}{!}{

\begin{tabular}{lccccccccc}
\toprule
\multirow{2}{*}{\textbf{Methods}} & \multicolumn{2}{c}{\textbf{Degree of Perturbation}(in \%)} & \multicolumn{2}{c}{\textbf{Contextual Detail}(in \%)} & \multicolumn{3}{c}{\textbf{Popularity}(in \%)} & \multirow{2}{*}{\textbf{Sensitivity Score}} \\
\cmidrule(lr){2-3} \cmidrule(lr){4-5} \cmidrule(lr){6-8}
& \textbf{rank 1} & \textbf{rank 2} & \textbf{rank 1} & \textbf{rank 2} & \textbf{rank 1} & \textbf{rank 2} & \textbf{rank 3}\\
\midrule

\multicolumn{9}{l}{\textbf{\uline{\textit{MusiQue $\bullet$ LLaMA-3-Instruct}}}} \\ [3pt] 
\addlinespace[3pt]
\rowcolor{lightblue}
Origin & 64.85 & 20.17 & 55.08 & 30.00 & 49.44 & 42.63 & 35.71 & 38.13 &\\
\rowcolor{lightblue}  
\textsc{Prompt} & 75.88 (\textcolor{magenta}{+11.03}) & 38.73 (\textcolor{magenta}{+18.56}) & 69.22 (\textcolor{magenta}{+14.14}) & 45.44 (\textcolor{magenta}{+15.44}) & 65.92 (\textcolor{magenta}{+16.48}) & 58.03 (\textcolor{magenta}{+15.40}) & 48.26 (\textcolor{magenta}{+12.55}) & 53.10 (\textcolor{magenta}{+14.97}) & \\
\rowcolor{lightblue}  
\rowcolor{lightblue} 
\textsc{CAD} & 62.10 (\textcolor{blue}{-2.65}) & 19.88 (\textcolor{blue}{-0.29}) & 51.69 (\textcolor{blue}{-3.39}) & 30.44 (\textcolor{magenta}{+0.44}) & 47.66 (\textcolor{blue}{-1.78}) & 40.62 (\textcolor{blue}{-2.01}) & 35.06 (\textcolor{blue}{-0.65}) & 37.04 (\textcolor{blue}{-1.09}) & \\
\rowcolor{lightblue}  
\textsc{COIECD} & 65.00 (\textcolor{magenta}{+0.15}) & 20.32 (\textcolor{magenta}{+0.32}) & 54.49 (\textcolor{blue}{-0.59}) & 30.88 (\textcolor{blue}{+0.88}) & 49.67 (\textcolor{magenta}{+0.23}) & 42.64 (\textcolor{magenta}{+0.01}) & 35.93 (\textcolor{magenta}{+0.22}) & 38.35 (\textcolor{magenta}{+0.22}) & \\
\rowcolor{lightblue}  
\textsc{CSKS w/o negative} & 69.41 (\textcolor{magenta}{+4.56}) & 44.18 (\textcolor{magenta}{+24.01}) & 67.74 (\textcolor{magenta}{+12.66}) & 45.88 (\textcolor{magenta}{+15.88}) & 61.69 (\textcolor{magenta}{+12.25}) & 54.46 (\textcolor{magenta}{+11.83}) & 54.32 (\textcolor{magenta}{+18.61}) & 53.96 (\textcolor{magenta}{+15.83}) & \\
\rowcolor{lightblue}  
\textsc{CSKS} & \textbf{78.08} (\textcolor{magenta}{+13.23}) & \textbf{60.38} (\textcolor{magenta}{+40.21}) & \textbf{79.97} (\textcolor{magenta}{24.89}) & \textbf{58.53} (\textcolor{magenta}{28.53}) & \textbf{75.27} (\textcolor{magenta}{+25.83}) & \textbf{65.84} (\textcolor{magenta}{+23.21}) & \textbf{66.66} (\textcolor{magenta}{+30.95}) & \textbf{66.72} (\textcolor{magenta}{+28.59}) & \\
\midrule

\multicolumn{9}{l}{\textbf{\uline{\textit{MusiQue $\bullet$ Qwen2.5-Instruct}}}} \\ [3pt] 
\addlinespace[3pt]
\rowcolor{lightpurple}  
Origin & 69.85 & 23.71 & 57.29 & 36.32 & 53.00 & 47.54 & 40.04 & 42.58 &\\
\rowcolor{lightpurple}  
\textsc{Prompt} & 76.76 (\textcolor{magenta}{+6.91}) & 36.08 (\textcolor{magenta}{+12.37}) & 67.60 (\textcolor{magenta}{+10.31}) & 45.29 (\textcolor{magenta}{+8.97}) & 62.81 (\textcolor{magenta}{+9.81}) & 58.48 (\textcolor{magenta}{+10.94}) & 48.27 (\textcolor{magenta}{+8.23}) & 52.32 (\textcolor{magenta}{+9.74}) & \\
\rowcolor{lightpurple}  
\rowcolor{lightpurple}  
\textsc{CAD} & 82.20 (\textcolor{magenta}{+12.35}) & 57.88 (\textcolor{magenta}{+34.17}) & 76.58 (\textcolor{magenta}{+19.29}) & 63.53 (\textcolor{magenta}{+27.21}) & 75.27 (\textcolor{magenta}{+22.27}) & 67.18 (\textcolor{magenta}{+19.64}) & 67.74 (\textcolor{magenta}{+27.70}) & 67.68 (\textcolor{magenta}{+25.20}) & \\
\rowcolor{lightpurple}  
\textsc{COIECD} & 69.85 (\textcolor{magenta}{+0.00}) & 24.74 (\textcolor{magenta}{+1.03}) & 57.58 (\textcolor{magenta}{+0.29}) & 37.06 (\textcolor{magenta}{+0.74}) & 53.45 (\textcolor{magenta}{+0.45}) & 47.54 (\textcolor{magenta}{+0.00}) & 41.13 (\textcolor{magenta}{+1.09}) & 43.21 (\textcolor{magenta}{+0.63}) & \\
\rowcolor{lightpurple}
\textsc{CSKS w/o negative} & 73.97 (\textcolor{magenta}{+4.12}) & 71.87 (\textcolor{magenta}{+48.16}) & 74.22 (\textcolor{magenta}{+16.93}) & 71.61 (\textcolor{magenta}{+35.29}) & 73.50 (\textcolor{magenta}{+20.50}) & 73.88 (\textcolor{magenta}{+26.34}) & 71.43 (\textcolor{magenta}{+31.39}) & 72.54 (\textcolor{magenta}{+29.96}) & \\
\rowcolor{lightpurple}  
\textsc{CSKS} & \textbf{94.85} (\textcolor{magenta}{+25.00}) & \textbf{85.13} (\textcolor{magenta}{+61.42}) & \textbf{90.43} (\textcolor{magenta}{+33.14}) & \textbf{89.56} (\textcolor{magenta}{+53.24}) & \textbf{93.54} (\textcolor{magenta}{+40.54}) & \textbf{85.94} (\textcolor{magenta}{+38.40}) & \textbf{90.47} (\textcolor{magenta}{+50.43}) & \textbf{89.26} (\textcolor{magenta}{+46.68}) & \\
\midrule

\multicolumn{9}{l}{\textbf{\uline{\textit{PopQA $\bullet$ LLaMA-3-Instruct}}}} \\ [3pt] 
\addlinespace[3pt]
\rowcolor{lightgreen}  
Origin & 52.04 & 23.62 & 52.21 & 23.48 & 43.14 & 37.29 & 33.22 & 34.32 &\\
\rowcolor{lightgreen}  
\textsc{Prompt} & 72.99 (\textcolor{magenta}{+20.95}) & 46.91 (\textcolor{magenta}{+23.29}) & 74.50 (\textcolor{magenta}{+22.29}) & 45.42 (\textcolor{magenta}{+21.94}) & 60.20 (\textcolor{magenta}{+17.06}) & 61.53 (\textcolor{magenta}{+24.24}) & 58.18 (\textcolor{magenta}{+24.96}) & 57.07 (\textcolor{magenta}{+22.75}) & \\
\rowcolor{lightgreen}  
\rowcolor{lightgreen}  
\textsc{CAD} & 47.63 (\textcolor{blue}{-4.41}) & 24.12 (\textcolor{magenta}{+0.50}) & 49.94 (\textcolor{blue}{-2.27}) & 21.85 (\textcolor{blue}{-1.63}) & 39.80 (\textcolor{blue}{-3.34}) & 36.85 (\textcolor{blue}{-0.44}) & 31.17 (\textcolor{blue}{-2.05}) & 32.69 (\textcolor{blue}{-1.63}) & \\
\rowcolor{lightgreen}  
\textsc{COIECD} & 53.03 (\textcolor{magenta}{+0.99}) & 23.62 (\textcolor{blue}{+0.00}) & 52.43 (\textcolor{magenta}{+0.22}) & 24.26 (\textcolor{magenta}{+0.78}) & 43.31 (\textcolor{magenta}{+0.17}) & 38.13 (\textcolor{magenta}{+0.84}) & 33.71 (\textcolor{magenta}{+0.49}) & 34.82 (\textcolor{magenta}{+0.50}) & \\
\rowcolor{lightgreen}
\textsc{CSKS w/o negative} & 59.64 (\textcolor{magenta}{+7.6}) & 53.09 (\textcolor{magenta}{+29.07}) & 67.99 (\textcolor{magenta}{+15.78}) & 43.77 (\textcolor{magenta}{+20.29}) & 56.18 (\textcolor{magenta}{+13.04}) & 57.52 (\textcolor{magenta}{+20.23}) & 53.97 (\textcolor{magenta}{+20.75}) & 54.13 (\textcolor{magenta}{+19.81}) & \\
\rowcolor{lightgreen}  
\textsc{CSKS} & \textbf{69.79} (\textcolor{magenta}{+17.75}) & \textbf{65.45} (\textcolor{magenta}{+41.83}) & \textbf{80.46} (\textcolor{magenta}{+28.25}) & \textbf{54.80} (\textcolor{magenta}{+31.32}) & \textbf{66.72} (\textcolor{magenta}{+23.58}) & \textbf{67.72} (\textcolor{magenta}{+30.43}) & \textbf{68.40} (\textcolor{magenta}{+35.18}) & \textbf{66.24} (\textcolor{magenta}{+31.92}) & \\
\midrule

\multicolumn{9}{l}{\textbf{\uline{\textit{PopQA $\bullet$ Qwen2.5-Instruct}}}} \\ [3pt] 
\addlinespace[3pt]
\rowcolor{lightpink}  
Origin & 66.15 & 28.59 & 60.60 & 34.18 & 51.67 & 47.83 & 42.79 & 43.59\\
\rowcolor{lightpink}  
\textsc{Prompt} & 75.63 (\textcolor{magenta}{+9.48}) & 40.17 (\textcolor{magenta}{+11.58}) & 71.85 (\textcolor{magenta}{+11.25}) & 43.99 (\textcolor{magenta}{+9.81}) & 58.86 (\textcolor{magenta}{+7.19}) & 57.86 (\textcolor{magenta}{+10.03}) & 57.05 (\textcolor{magenta}{+14.26}) & 54.63 (\textcolor{magenta}{+11.04}) & \\
\rowcolor{lightpink}  
\rowcolor{lightpink}  
\textsc{CAD} & 78.06 (\textcolor{magenta}{+11.91}) & 61.15 (\textcolor{magenta}{+32.56}) & 78.04 (\textcolor{magenta}{+17.44}) & 61.19 (\textcolor{magenta}{+27.01}) & 70.73 (\textcolor{magenta}{+19.06}) & 69.23 (\textcolor{magenta}{+21.40}) & 68.88 (\textcolor{magenta}{+26.09}) & 67.80 (\textcolor{magenta}{+24.21}) & \\
\rowcolor{lightpink}  
\textsc{COIECD} & 65.82 (\textcolor{blue}{-0.33}) & 28.04 (\textcolor{blue}{-0.55}) & 59.49 (\textcolor{blue}{-1.11}) & 34.40 (\textcolor{magenta}{+0.22}) & 50.50 (\textcolor{blue}{-1.17}) & 47.32 (\textcolor{blue}{-0.51}) & 43.11 (\textcolor{magenta}{+0.32}) & 43.31 (\textcolor{blue}{-0.28}) & \\
\rowcolor{lightpink}
\textsc{CSKS w/o negative} & 68.02 (\textcolor{magenta}{+1.87}) & 75.83 (\textcolor{magenta}{+47.24}) & 74.06 (\textcolor{magenta}{+13.46}) & 69.79 (\textcolor{magenta}{+35.61}) & 75.08 (\textcolor{magenta}{+23.41}) & 70.57 (\textcolor{magenta}{+22.74}) & 70.17 (\textcolor{magenta}{+27.38}) & 71.77 (\textcolor{magenta}{+28.18}) & \\
\rowcolor{lightpink}  
\textsc{CSKS} & \textbf{93.83} (\textcolor{magenta}{+27.68}) & \textbf{90.40} (\textcolor{magenta}{+61.81}) & \textbf{93.27} (\textcolor{magenta}{+32.67}) & \textbf{90.96} (\textcolor{magenta}{+56.78}) & \textbf{88.46} (\textcolor{magenta}{+36.79}) & \textbf{93.14} (\textcolor{magenta}{+45.31}) & \textbf{94.65} (\textcolor{magenta}{+51.86}) & \textbf{92.24} (\textcolor{magenta}{+48.65}) & \\
\bottomrule
\end{tabular}
}
\caption{Accuracy when evaluated on specific ranks of individual dimensions in the dataset and the overall \textit{Sensitivity Score}. For each dimension, Rank 1 represents the least challenging cases, while higher ranks indicate increasing difficulty. \ourmethod{} outperforms baseline methods under all metrics.}
\vspace{-4mm}
\label{tab:main-results}
\end{table*}

\section{Experiments}

\subsection{Baselines}
\label{subsection:baselines}
We adopt representative baselines of three types, specifically, prompting, decoding-time strategy (CAD \citep{shi-etal-2024-trusting}, COIECD \citep{yuan-etal-2024-discerning}), and neuron-editing method (IRCAN \citep{shi2024ircan}). The baselines' details and relevant configurations are in Appendix \ref{appendix:baselines}.

Besides, since the positive model $\mathcal{P}$ is already fine-tuned to adhere to the context, its distribution score $\mathcal{D}_\mathcal{P}$ can amplify the importance of contextual information. Thus, it's natural to ask whether it's necessary to use another negative model. For this purpose, we replace the negative model with the original small model and refer to this configuration as "CSKS w/o negative".
\subsection{Models and Settings}
We employ two state-of-the-art instruction-tuned LLMs as target models: Llama-3-70B-Instruct \citep{dubey2024llama} and Qwen2.5-72B-Instruct \citep{yang2024qwen2}\footnote{To illustrate transferability, we further expand our experiment on another model family, gemma-2-27b-it \citep{team2024gemma} and show the results in Appendix \ref{appendix:transfer}}. For each target model, we utilize its smaller counterpart as proxy model – specifically, fine-tuned versions of Llama-3-8B-Instruct for the Llama-3 series and Qwen2.5-7B-Instruct for the Qwen2.5 series. We use greedy decoding in all the experiments to ensure reproducibility.

For constructing the evaluation dataset, we use MuSiQue \citep{trivedi-etal-2022-musique} and PopQA \cite{mallen-etal-2023-trust}, both widely used question-answering datasets, as the source datasets. Following the setup in \citep{shi2024ircan}, we format the task as binary-choice questions. Correct options correspond to the answers in context, and the incorrect options correspond to the original answers to the question. This design creates controlled knowledge conflict scenarios where model performance directly reflects its ability to prioritize contextual or parametric knowledge. It is important to clarify that the contextual answers used here are exactly the perturbed answers we introduce during dataset construction.

To comprehensively evaluate the model's performance across the entire dataset, we use accuracy as a default metric, calculated per rank within our three operational dimensions (perturbation, context, popularity). Additionally, we employ the previously defined \textit{Sensitivity Score} to assess the model's ability to adhere to the given context, which is also normalized into a 100-scale.

\subsection{Results}

Table \ref{tab:main-results} shows \ourmethod{} consistently outperforms all baselines across all evaluation dimensions, achieving substantial average sensitivity score improvements (LLaMA-3: +30.26, Qwen2.5: +47.67). Key observations include:
\begin{enumerate}
    
    \item \textbf{Baseline Limitations}: Decoding-time strategy baselines exhibit inconsistent effectiveness. While \textsc{CAD} shows moderate gains on Qwen2.5 (+24.2 sensitivity score), it degrades performance on LLaMA-3 (-1.1 sensitivity score). \textsc{COIECD}'s entropy-based constraints seem insufficient for resolving deep parametric conflicts, yielding marginal improvements of less than 1.5 across all configurations. The core reason for these limitations lies in how the steering signal is generated. \textsc{CAD} and \textsc{COIECD} rely on a single model's self-contrast (with vs. without context). However, as large models are "stubborn" and resistant to deviating from their strong parametric knowledge, this "self-guidance" signal is often too weak to overcome the model's own biases. 

    In contrast, \ourmethod{} derives a powerful and explicit steering signal from the difference between two smaller, more adaptable proxy models that have been contrastively tuned. This use of external, specialized "experts" provides a much stronger and more reliable guide towards contextual faithfulness, explaining its significantly superior performance.
    
    \item \textbf{Robustness of \ourmethod{} and the Synergy with Negative Models}: The "CSKS w/o negative" configuration (replacing the negative model with the original small model) remains competitive, outperforming other baselines (e.g., +15.83 sensitivity for LLaMA-3 in MusiQue). This indicates the robustness of the core \ourmethod{} framework, as it can leverage the proxy model's knowledge to mitigate parametric conflicts even without explicit negative sampling. This finding also hints at potential cost-saving opportunities in practical implementations. On the other hand, incorporating the negative model further boosts the performance (MusiQue avg. sensitivity: LLaMA-3 +28.59, Qwen2.5 +46.68), highlighting its critical role in enhancing the framework's ability to distinguish between contextual and intrinsic knowledge.
    
    \item \textbf{Dimensional Sensitivity}: Among the three dimensions we introduce, the perturbation degree has the greatest effect: large perturbations create obvious conflicts demanding resolution, while small, subtle deviations are more confounding, making it harder for the model to choose between external context and internal knowledge. Furthermore, \ourmethod{} smooths or even reverses differences across popularity ranks, indicating its efficacy in mitigating pre-training bias associated with entity popularity.
\end{enumerate}


\begin{figure*}[t]
    \centering
    \includegraphics[width=\textwidth]{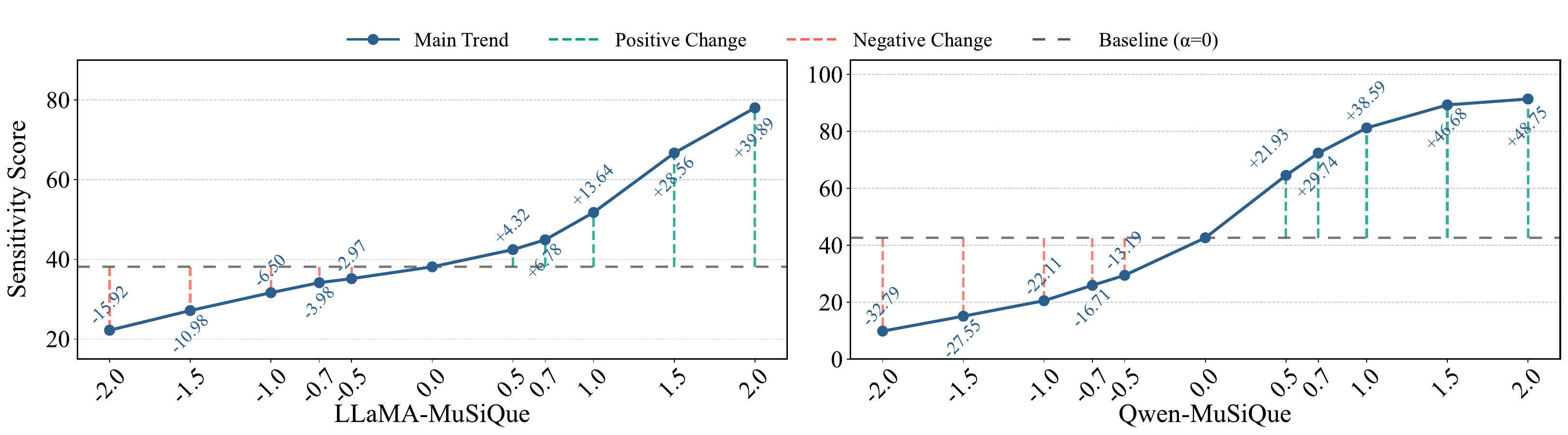}  
    \caption{The performance of LLaMA and Qwen controlled bidirectionally, demonstrating the continuous adjustment capability of our method from two directions.}
    \label{fig:trend}
    \vspace{-5mm}
\end{figure*}

After showing the effectiveness of \ourmethod{} framework, we further show that our framework can achieve continuous and precise control over the knowledge sensitivity to contextual knowledge through the steering parameter $\alpha$. As illustrated in Figure \ref{fig:trend}, increasing $\alpha$ values ($\alpha > 0$) produce a monotonic enhancement of sensitivity score from 4.32 to 39.80 for LLaMA on MuSiQue, with potential for further increase. This directional control proves critical for applications requiring dynamic knowledge updates, where models must suppress outdated parametric knowledge in favor of fresh contextual evidence. Results on PopQA can be found in Appendix \ref{appendix:popqa}.)

The previous experiments demonstrate the effectiveness of \ourmethod{} framework when aggregating new and conflicting knowledge in context setting $\alpha > 0$. Notably, extending $\alpha$ to negative values ($\alpha < 0$) reveals an inverse mode of action: the framework can suppress contextual influence to amplify parametric reliance. As demonstrated in Figure \ref{fig:trend}, setting $\alpha=-2.0$ reduces contextual sensitivity score by $15.9$ for LLaMA and $32.8$ for Qwen compared to their baselines ($\alpha = 0$), effectively transforming the target model into a parametric knowledge conservative. This bidirectional control mechanism ($\alpha \in (-\infty, +\infty)$) enables continuous scenario adaptation, allowing practitioners to calibrate models for either context-sensitive scenarios or parametric knowledge preservation.
\vspace{-1mm}
\begin{figure}[t]
\centering
\begin{minipage}{0.50\textwidth}
    \centering
    \includegraphics[width=\textwidth]{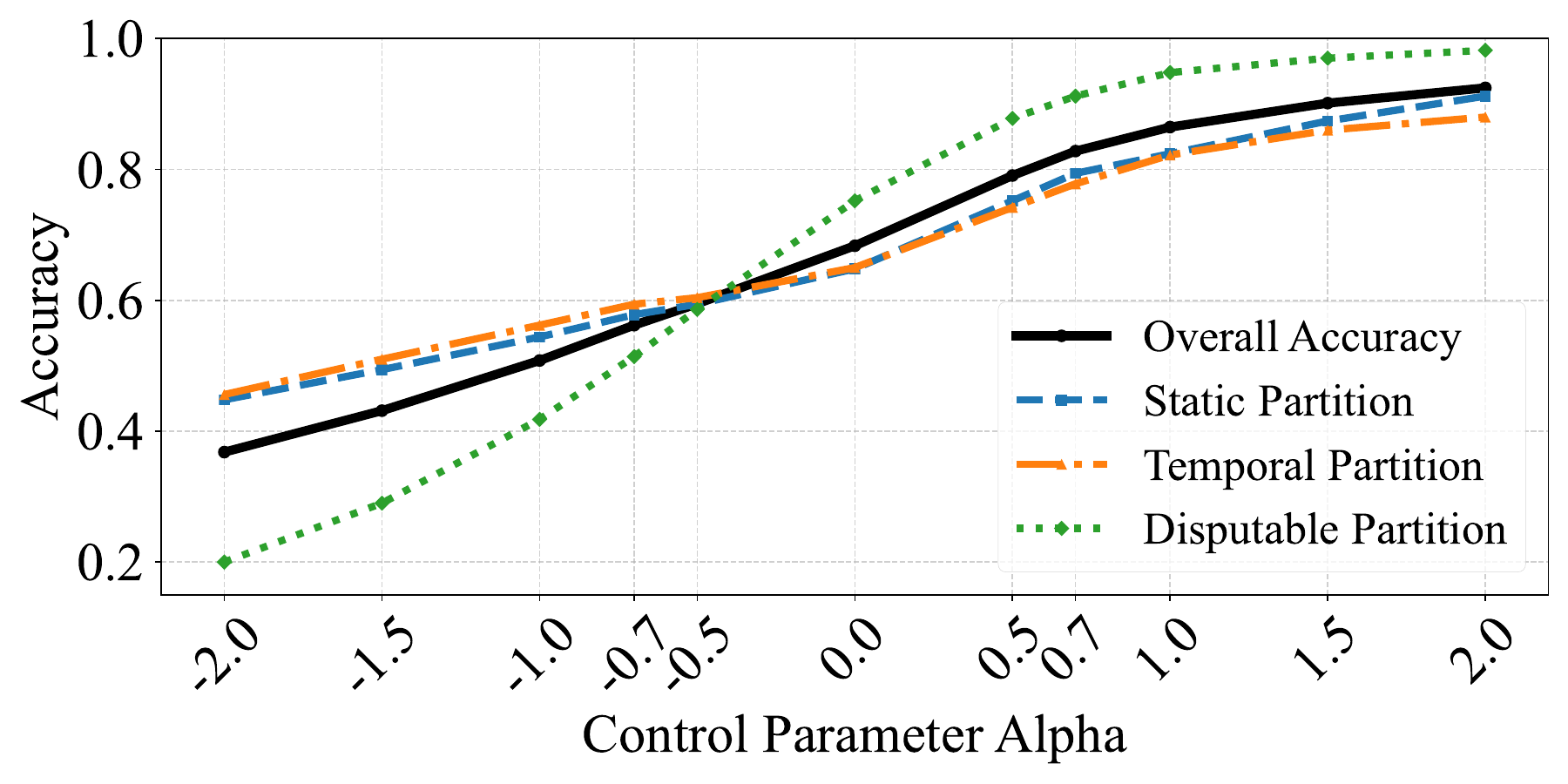}  
    \vspace{-6mm}
    \caption{Accuracy of Qwen2.5 steered by \ourmethod{} on the DynamicQA dataset as a function of the control parameter $\alpha$. Results are shown overall and broken down by conflict partition type, demonstrating \ourmethod{}'s effectiveness and continuous controllability in handling diverse real-world knowledge conflicts.}
    \label{fig:dynamicQA}
\end{minipage}
\vspace{-6mm}
\end{figure}

\subsection{Real-World Knowledge Conflicts Evaluation}

To address concerns about the reliance on synthetic datasets and further validate the practical applicability of \ourmethod{}, we conducted an additional experiment on the DynamicQA benchmark \citep{marjanovic-etal-2024-dynamicqa}. DynamicQA is designed to evaluate LLMs' ability to handle knowledge conflicts arising from evolving real-world information. It categorizes questions based on conflict types: \textbf{Static} (there is only one possible representation of such facts), \textbf{Temporal} (conflicts arising from knowledge updated over time) and \textbf{Disputable} (conflicts where reliable sources disagree). This setting allows us to assess \ourmethod{}'s performance in more realistic and diverse conflict scenarios. 

We provide the results of Qwen2.5-72B-Instruct steered by \ourmethod{} on DynamicQA in Figure \ref{fig:dynamicQA}, with varied control parameter $\alpha$ from -2.0 to +2.0. The accuracy was measured separately for each conflict partition type, as well as overall. We also provide results of other methods (Origin, Prompt, CAD and COIECD) and their comparison with \ourmethod{} in Appendix \ref{appendix:dynamicQA}. Consistent with our findings on synthetic datasets, \ourmethod{} demonstrates continuous control over the model's contextual sensitivity. As $\alpha$ increases, the overall accuracy monotonically improves, indicating enhanced faithfulness to the provided context.


\begin{table}[t]
    \centering
    \resizebox{\linewidth}{!}{ 
        \begin{tabular}{c|ccccc}
            \toprule[1.5pt]
            \textbf{Alpha} & \textbf{STEM} & \textbf{Humanities} & \textbf{Other} & \textbf{Social} & \textbf{Average} \\
           \midrule[1pt]
            -2.0  & 89.34  & 78.01  & 88.27  & 82.54  & 85.00  \\
            -1.5  & 90.98  & 77.66  & 88.08  & 83.81  & 85.44  \\
            -1.0  & 91.39  & 77.32  & 88.64  & 83.17  & 85.51  \\
            -0.7  & 91.39  & 78.69  & 88.64  & 84.13  & 86.01  \\
            -0.5  & 91.39  & 79.73  & 89.01  & 84.44  & 86.45  \\
            \rowcolor{lightgray}
            \textbf{72B($\alpha=0$)}  & \textbf{92.62}  & \textbf{79.04}  & \textbf{88.64}  & \textbf{84.76}  & \textbf{86.45}  \\
            +0.5  & 91.80  & 78.01  & 87.71  & 84.44  & 85.65  \\
            +0.7  & 91.80  & 78.69  & 87.52  & 84.13  & 85.65  \\
            +1.0  & 90.98  & 78.01  & 87.34  & 83.81  & 85.22  \\
            +1.5  & 90.98  & 76.29  & 85.85  & 83.49  & 84.21  \\
            +2.0  & 90.98  & 74.91  & 84.92  & 81.27  & 83.06  \\
            \midrule[1pt]
            \rowcolor{lightgray}  
            \textbf{7B} & \textbf{84.84} & \textbf{70.79} & \textbf{76.35} & \textbf{76.83} & \textbf{76.78} \\
            \bottomrule[1.5pt]
        \end{tabular}
    }
    \caption{Performance comparison showing trade-off between faithfulness to contextual knowledge and general capabilities.}
    \vspace{-6mm}
    \label{tab:general_ability_results}
\end{table}
\vspace{-1mm}
\subsection{Analysis}


\begin{figure*}[ht]
    \centering
    \includegraphics[width=\textwidth]{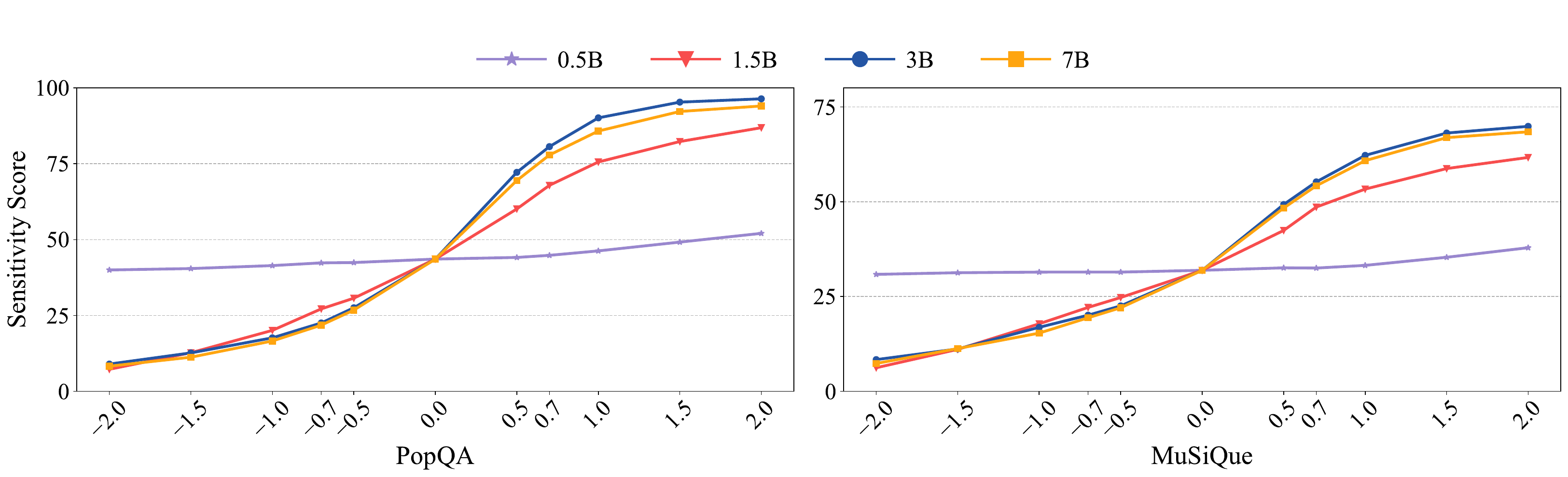}  
    \caption{The performance of \ourmethod{} under varying proxy model sizes on MuSiQue and PopQA respectively. Smaller proxy models (0.5B, 1.5B) have a marginal yet increasing effect on the 72B target model's sensitivity score. Remarkably, the 3B proxy model matches the 7B model in sensitivity adjustment, validating that our framework enables potent context sensitivity modulation using substantially smaller models.}
    \vspace{-2mm}
    \label{fig:size}
\end{figure*}

\paragraph{The Impact of Proxy Model Size}
To explore resource savings with smaller proxy models, we use the Qwen2.5 family (0.5B to 7B) to steer a 72B model under our framework. As shown in the Figure \ref{fig:size}, the 0.5B proxy has a subtle but growing impact on the target model's sensitivity score, while the 1.5B proxy's impact already becomes very significant. A 3B proxy's impact is comparable, occasionally slightly better, than a 7B proxy. These results demonstrate our framework can adjust context sensitivity on a much larger model with significantly smaller overhead (e.g., using a 3B proxy). This efficiency may stem from our framework's \textit{selective steering} mechanism, where proxy models focus exclusively on context sensitivity modulation rather than full knowledge representation.

\paragraph{Trade-Off Discussion}  To study how scaling the control parameter $\alpha$ would impact the general capabilities of the model, we conduct an evaluation on the MMLU benchmark \citep{hendryckstest2021} for world knowledge understanding ability of LLMs (complex reasoning on 2WikiMultiHopQA \citep{ho-etal-2020-constructing} is detailed in Appendix \ref{appendix:reasoning}). For simplicity, we tested on two tasks from each of MMLU's four subjects (STEM, Humanities, Social, and Other). The experiment results in Table \ref{tab:general_ability_results} reveal a crucial trade-off in knowledge sensitivity control: while increasing the absolute value of $\alpha$ enables extensive adjustment of the model's contextual sensitivity (Figure \ref{fig:trend}), excessive values ($|\alpha|>1.5$) lead to noticeable degradation in general capabilities, particularly Humanities (-4.10\%) domain. This performance decline suggests that extreme sensitivity adjustments may disrupt the target model's fundamental reasoning patterns, highlighting the importance of maintaining a balanced $\alpha$ range that preserves core competencies while enabling effective knowledge adaptation. Notably, even with substantial $\alpha$ variation, the target 72B model consistently outperforms the 7B model by significant margins (average +8.67\%), demonstrating our framework successfully leverages the large model's superior general ability alongside precise sensitivity control. These findings indicate that strategic $\alpha$ selection can achieve an effective equilibrium between contextual adaptability and general capability preservation, fulfilling our framework's dual objectives of precise knowledge steering and performance maintenance.
\vspace{-1mm}
\begin{table}[t]
    \centering
    \resizebox{\linewidth}{!}{ 
        \begin{tabular}{ccccccc}
            \toprule
            \textbf{Raw} & \textbf{$\alpha=0.5$} & \textbf{$\alpha=0.7$} & \textbf{$\alpha=1.0$} & \textbf{$\alpha=1.5$} & \textbf{$\alpha=2.0$}\\
            \midrule
            \multicolumn{5}{l}{\textbf{\uline{\textit{MusiQue $\bullet$ Proxy-LLaMA}}}} \\ [3pt] 
            \addlinespace[3pt]
            \cellcolor{red!10} 51.24 & \cellcolor{green!10} 60.38& \cellcolor{green!10} 66.36 & \cellcolor{green!10}76.32 & \cellcolor{green!10}87.79 & \cellcolor{green!10} 93.45\\
            \midrule
            \multicolumn{5}{l}{\textbf{\uline{\textit{PopQA $\bullet$ Proxy-Qwen}}}} \\ [3pt] 
            \addlinespace[3pt]
            \cellcolor{red!10} 56.56 & \cellcolor{green!10} 75.07& \cellcolor{green!10} 84.67 & \cellcolor{green!10}90.89 & \cellcolor{green!10}93.58 & \cellcolor{green!10} 94.73\\
            \bottomrule
        \end{tabular}
    }
    \caption{Performance of GPT-3.5-Turbo steered by LLaMA and Qwen. Our method also works for black-box models such as GPT-3.5-Turbo.}
    \vspace{-2mm}
    \label{tab:black-box}
\end{table}

\paragraph{Extending to Black Box Model} For the black-box models that we can't obtain weights, our framework remains effective. We apply our framework to adapt GPT-3.5-Turbo \citep{ouyang2022training}. In this setting, since we can only access the log probabilities for the top five tokens through the API, \ourmethod{} only reweights the five tokens. We present the results in Table \ref{tab:black-box}. For black-box models that do not belong to the same model family as the proxy model, \ourmethod{} can still effectively control its context sensitivity, demonstrating its broad application domain.

\vspace{-1mm}
\section{Related Works}

\subsection{Knowledge Conflicts}
\vspace{-1mm}
Knowledge conflicts occur when contextual knowledge contradicts parametric knowledge \citep{mallen-etal-2023-trust, xu-etal-2024-knowledge-conflicts,kortukov2024studying}. Previous research often prioritized contextual knowledge over parametric knowledge for LLM responses \citep{gekhman2023trueteacher, lee-etal-2022-plug, shi2024incontext, zhang-etal-2020-dialogpt,zhou-etal-2023-context}. This is a valuable setting for applications such as retrieval-augmented LMs \citep{ram-etal-2023-context, shi-etal-2024-replug}, where the context may be of high quality (e.g., containing updated knowledge). However, varying context quality across scenarios means that a constant reliance on context is insufficient—an underexplored issue. We advocate for precise, continuous control over LLMs' contextual reliance and propose an effective, efficient framework to achieve this. Another line of work focuses on evaluating and understanding LLMs in knowledge conflicts and mining factors affecting LLMs' choice in knowledge conflicts. For instance, contextual detail affects LLM choices \citep{Wu2024ClashEvalQT, tan-etal-2024-blinded}; LLMs favor popular entity information and are sensitive to data presentation order \citep{Xie2023AdaptiveCO}; models resist obviously false permuted knowledge \citep{qian2024merge}; and increased conflicting hops challenge LLM reasoning \citep{jin2024tug}. We leverage these key factors to measure knowledge manipulation difficulty and offer a more comprehensive evaluation method. We further utilize the key factors to measure the difficulty of manipulating certain knowledge and provide a more comprehensive evaluation method.
\vspace{-1mm}
\subsection{Updating Knowledge in Language Models}
To introduce new knowledge to LMs, previous works explore tuning-based approaches \citep{wang2024resolving}, decoding strategies \citep{shi-etal-2024-trusting, zhao2024steeringknowledgeselectionbehaviours, wang2024adacad}, and model editing methods \citep{meng2023massediting, gupta-etal-2023-editing,shi2024ircan}. Nevertheless, these methods are usually inefficient or ineffective for large models, not workable for black-box models, or unable to continuously adjust LLMs’ sensitivity to the new contextual knowledge, while our approach can steer LLMs' sensitivity to contextual knowledge continuously at a lightweight cost.
\subsection{Control of Language Models}
Motivated by LMs' growing capabilities \citep{li-etal-2023-defining}, many studies focus on controlling certain attributes of LM generation, usually non-toxicity and positive sentiment. Representation engineering is a common solution. \citet{han-etal-2024-word} use word embeddings to steer LMs for language model detoxification and sentiment control. \citet{zhao2024steeringknowledgeselectionbehaviours} steer knowledge behaviors of LLMs with SAE-based representation engineering. \citet{zeng-etal-2025-towards} and \citet{tan2024knowledge} leverage LLMs’ internal representations for knowledge integration and security. Some other works tune the hidden representations of LMs to change behaviors \citep{wu2024reft, hernandez2024inspecting, li2023inferencetime,4oapi}. Another line of work incorporates other models to guide the generation process \citep{liu-etal-2021-dexperts, liu2024tuning, feng-etal-2024-modular}. Our work also borrows this idea but emphasizes controlling sensitivity to contextual knowledge and achieves precise and continuous control.

\section{Conlusion}
We present \ourmethod{}, an efficient and effective framework using small LMs as proxies to adjust output distributions of LLMs, thus controlling LLMs' sensitivity to knowledge provided in context. We also introduce a fine-grained evaluation method for this sensitivity. Extensive experiments demonstrate that our framework achieves state-of-the-art, more crucially, achieves precise and continuous control over how LLMs utilize information from context.

\section*{Limitations}
While we show \ourmethod{}'s effective control of LLMs in knowledge adaptation, the optimal calibration of the guiding hyperparameter $\alpha$ may vary in real scenarios where a balance between knowledge adaptation and LLMs' general abilities is essential.  Future research could further explore methods for automatically or more adaptively determining the value of $\alpha$ to enhance the practical flexibility of the \ourmethod{} framework.

\section*{Acknowledgement}
Yilin Wang ran all the experiments in the paper and drafted part of the paper. Heng Wang designed the conceptual framework, designed the experiments with Yilin Wang, and drafted part of the paper. Yuyang Bai provided feedback on the paper draft. Minnan Luo provided feedback throughout the project.

We thank Shangbin Feng for providing constructive feedback on our project. We would also like to thank all LUD Lab members for our collaborative research environment. This work is supported by the National Nature Science Foundation of China (No. 62192781, No. 62272374), the Natural Science Foundation of Shaanxi Province (No. 2024JC-JCQN-62), the Key Research and Development Project in Shaanxi Province (No. 2023GXLH-024), the Project of China Knowledge Center for Engineering Science and Technology, the Project of Chinese Academy of Engineering “The Online and Offline Mixed Educational Service System for ‘The Belt and Road’ Training in MOOC China”, and the K. C. Wong Education Foundation.


\clearpage

\appendix

\label{sec:appendix}

\section{Finetune Dataset Construction Details}
\label{appendix:finetune dataset}
\begin{figure}[t]
\centering
\begin{minipage}{0.5\textwidth} 
    \centering
    \includegraphics[width=\textwidth]{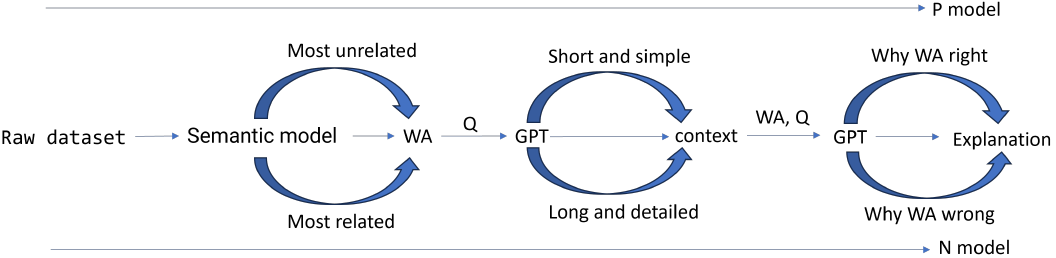}  
    \caption{The pipeline to get the data used to finetune our $\mathcal{P}$ model and $\mathcal{N}$ model}
    \label{fig:finetune-detail}
\end{minipage}
\end{figure}

To obtain our $\mathcal{P}$ model and $\mathcal{N}$ model, we fine-tune the Llama-3-8B-instruct model and Qwen-2.5-7B-instruct model. To ensure generalization, the fine-tuning datasets are constructed using methods and domains \textbf{different} from those of the synthesized conflict datasets in our main experiment. To achieve optimal results, we have designed a specialized pipeline for constructing the fine-tuning dataset as shown in Figure \ref{fig:finetune-detail}.

We select ECQA as the base dataset, which is a multiple-choice QA dataset where each question is accompanied by five answer options.

\begin{itemize}
    \item For the $\mathcal{P}$ model: We select the incorrect option least related to the correct answer as the "contextual answer."
    \item For the $\mathcal{N}$ model: We select the incorrect option most related to the correct answer as the "contextual answer."
\end{itemize}

Next, using GPT, we generate supportive context based on the chosen answer and the question. 

\begin{itemize}
    \item For the $\mathcal{P}$ model, the generated context was short and simple.
    \item For the $\mathcal{N}$ model, the context was long and detailed.
\end{itemize}

Finally, we again use GPT to generate explanations based on the context, question, and selected answer.

\begin{itemize}
    \item For the $\mathcal{P}$ model, the explanation justified why the selected answer was correct.
    \item For the $\mathcal{N}$ model, the explanation detailed why the selected answer was incorrect.
\end{itemize}

Using these constructed answers and their corresponding explanations, we fine-tune the model as follows:

\begin{itemize}
    \item The $\mathcal{P}$ model was fine-tuned on the selected answers and their associated explanations.
    \item The $\mathcal{N}$ model was fine-tuned on the original correct answers and their explanations.
\end{itemize}

\section{Effectiveness of the Grading System}

\begin{figure}[t]
\centering
\begin{minipage}{0.5\textwidth} 
    \centering
    \includegraphics[width=\textwidth]{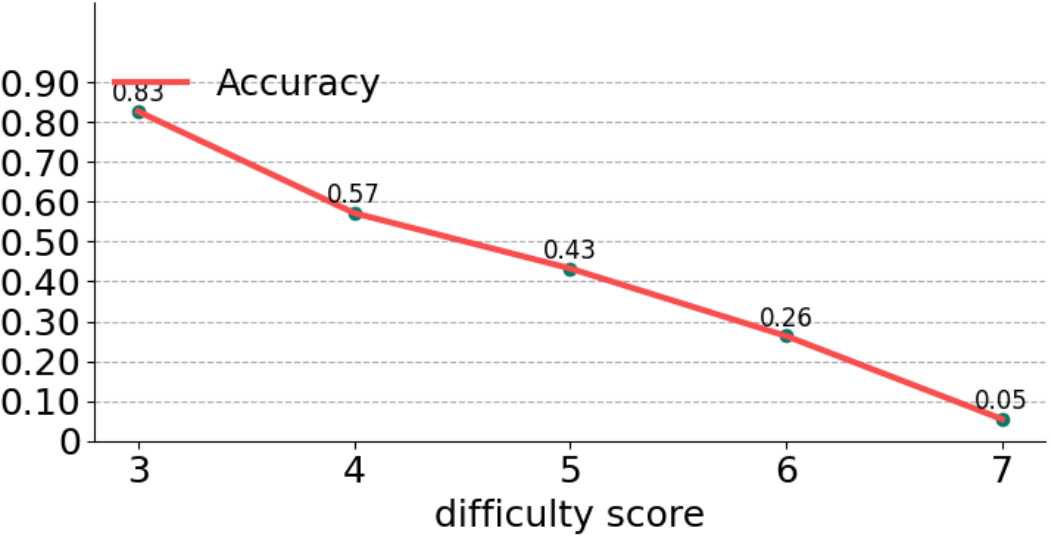}  
    \caption{The accuracy of the LLaMA-3-70B-Instruct model across questions of each difficulty score.}
    \label{fig:effectiveness}
\end{minipage}
\end{figure}

To validate the effectiveness of our grading system, we conduct a validation experiment. We analyze the accuracy of the target model across questions of varying difficulty levels, with the results shown in Figure \ref{fig:effectiveness}. The results reveal that as question difficulty increases, accuracy correspondingly decreases. This demonstrates that our grading system successfully quantifies problem difficulty.

\section{Baselines}
\label{appendix:baselines}

The baselines we adopt in our main experiment are:
\begin{itemize}
    \item \textbf{Origin}: refers to naive LLMs without any modifications.
    \item \textbf{Prompt}: prompts LLMs with explicit instructions to ensure their answers align with the given context.
    \item \textbf{IRCAN} : identifies context-responsive neurons within the LLM's feedforward network (FFN) layers and enhances their activation to improve the utilization of contextual information.
    \item \textbf{CAD} : is a decoding-time strategy that adjusts the output probabilities of LLMs to emphasize differences between context-aware and context-agnostic scenarios.
    \item  \textbf{COIECD} : adapts its decoding strategy based on a contextual information-entropy constraint to discern when a context generates conflicting knowledge with the model's internal knowledge.
\end{itemize}
For CAD and COIECD, we use the optimal hype-parameters reported in their papers for baselines. For our method, we do not search for an optimal parameter but just setting $\alpha$ the to same as CAD.
To check whether these baselines are effective, we conducted a verification on small model. The results are presented in Appendix \ref{appendix:small modls}, which shows that while all baseline methods work fine for the small model, IRCAN shows minimal performance enhancement. This limited efficacy combined with IRCAN's significantly larger computational overhead makes it unsuitable for our primary objective of efficient large-model adaption. So we exclude IRCAN from our main experiments.

\section{Fine-tune results on small models}
\label{appendix:small modls}
\begin{figure}[th]
\centering
\begin{minipage}{0.5\textwidth} 
    \centering
    \includegraphics[width=\textwidth]{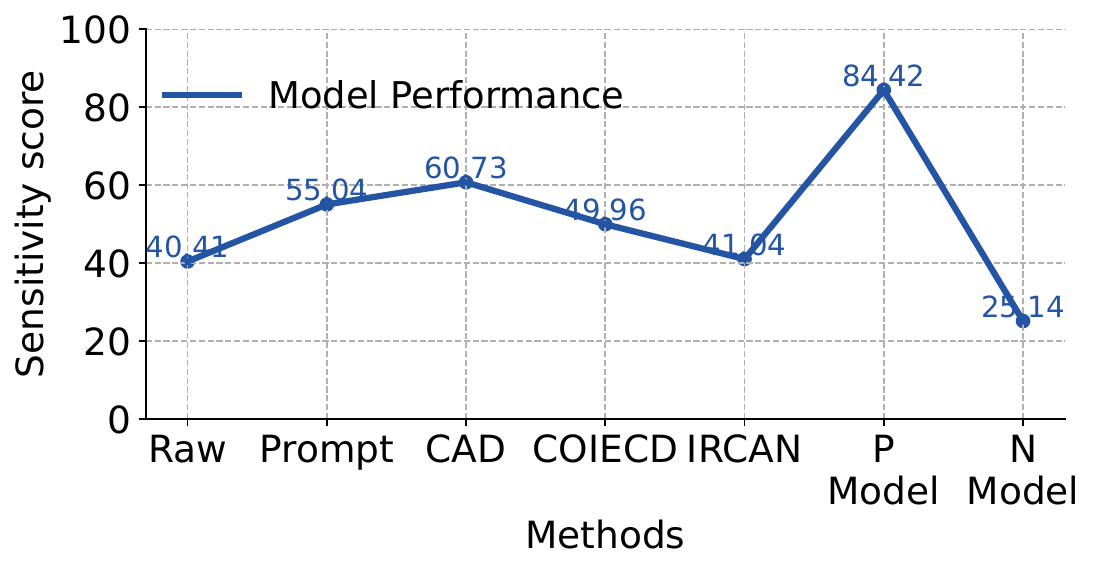}  
    \caption{The effects of different methods on the LLaMA-3-8B-instruct model tested on PopQA.}
    \label{fig:8B-results}
\end{minipage}
\end{figure}

Figure \ref{fig:8B-results} illustrates the effects of different methods on the LLaMA-3-8B-instruct model. From the results, we observe the following:

\begin{enumerate}
    \item The Prompt, CAD and COIECD methods all improve the performance of the 8B small model, while the impact of IRCAN on the small model's performance is minimal.
    \item We also present the performance of our fine-tuned $\mathcal{P}$ model and $\mathcal{N}$ model. The $\mathcal{P}$ model performs the best, as it effectively incorporates knowledge from the context, while the $\mathcal{N}$ model scores much lower, indicating that it tends to rely on its internal knowledge and resists external contextual information. This indicates that our fine-tuning is successful.
\end{enumerate}

\section{Steering Results on PopQA}
\label{appendix:popqa}
We present the steering results on the PopQA dataset, which have similar trend as that on the MuSiQue dataset.
\begin{figure}[h!]
    \centering
    \begin{subfigure}[b]{0.45\textwidth}
        \centering
        \includegraphics[width=\textwidth]{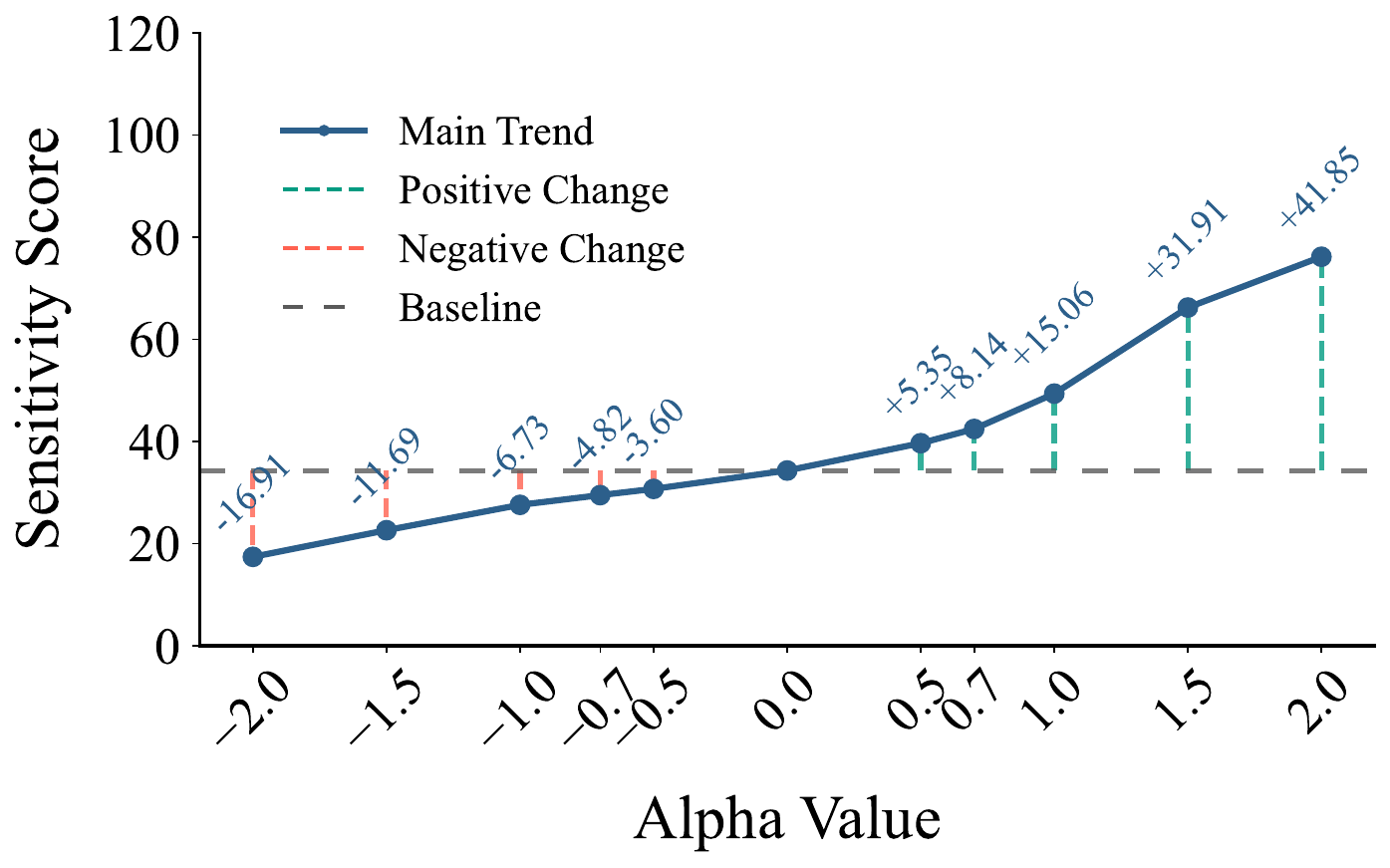}
        \caption{Sensitivity Score Variation with Alpha Values on LLaMA-PopQA}
        \label{fig:llama-popqa}
    \end{subfigure}
    \hspace{0.5cm}
    \begin{subfigure}[b]{0.45\textwidth}
        \centering
        \includegraphics[width=\textwidth]{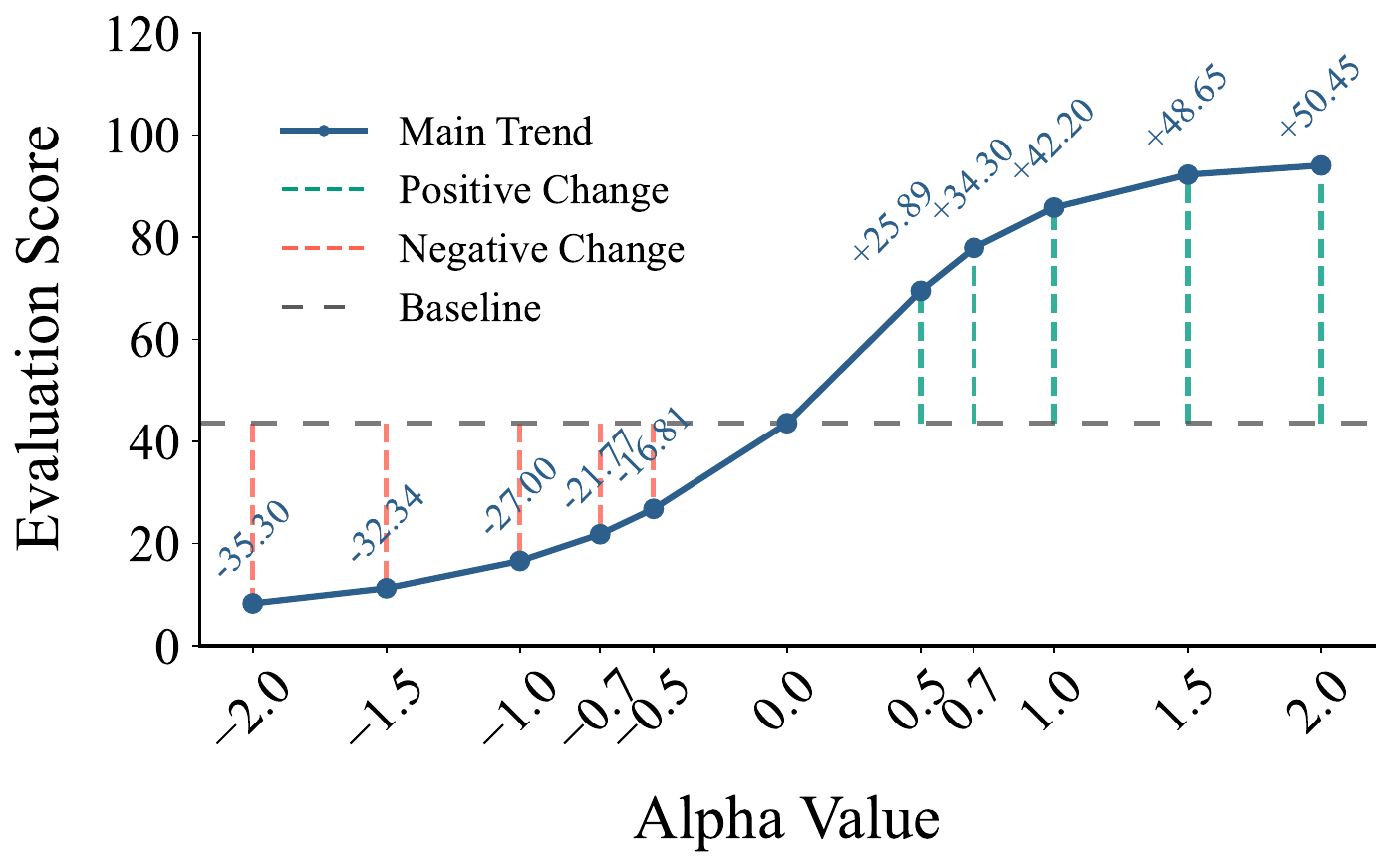}
        \caption{Sensitivity Score Variation with Alpha Values on Qwen-PopQA}
        \label{fig:qwen-popqa}
    \end{subfigure}
    \caption{Sensitivity score variation with alpha values on PopQA.}
    \label{fig:sensitivity}
\end{figure}

\section{Performance Comparison on DynamicQA}
\label{appendix:dynamicQA}

\begin{figure}[t]
\centering
\begin{minipage}{0.5\textwidth} 
    \centering
    \includegraphics[width=\textwidth]{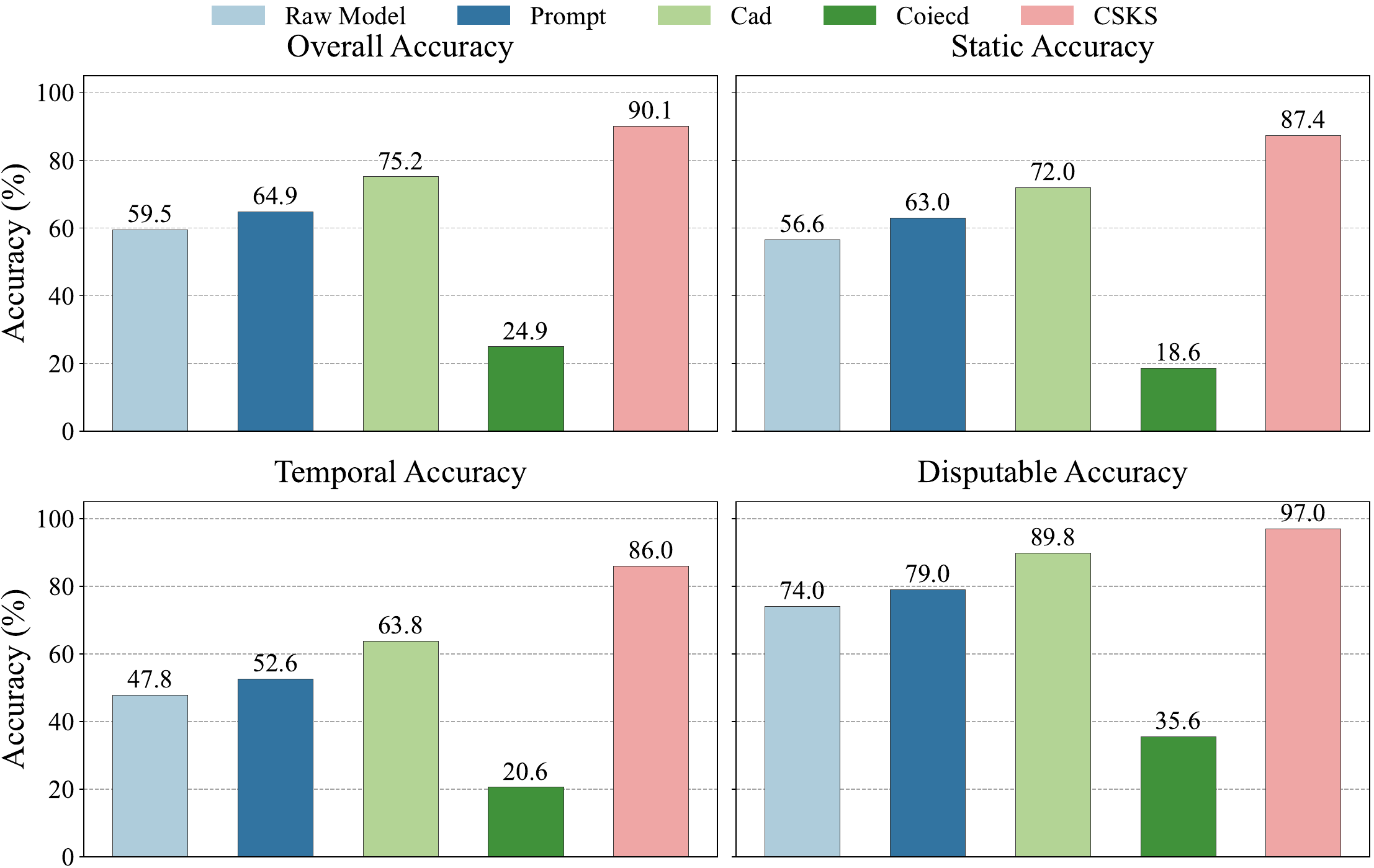}  
    \caption{Comparative performance (Accuracy \%) of \ourmethod{} and baseline methods (Raw Model, Prompt, CAD, COIECD) on the DynamicQA dataset. Results are shown for Overall Accuracy and broken down by conflict partition types: Static, Temporal, and Disputable. \ourmethod{} consistently outperforms all baseline methods across all categories.}
    \label{fig:dynamicqa-all}
\end{minipage}
\end{figure}

Figure \ref{fig:dynamicqa-all} presents a head-to-head comparison of these methods across overall accuracy and specific conflict partition types (Static, Temporal, and Disputable) on DynamicQA. Across all evaluated dimensions, \ourmethod{} consistently and substantially outperforms all baseline approaches.

The consistently superior performance of \ourmethod{} across diverse real-world conflict types underscores its robustness and practical advantages over existing methods for managing knowledge conflicts in LLMs. The substantial margins, especially in the more challenging Disputable partition, further validate the efficacy of our proxy-based steering mechanism.

\section{The \ourmethod{} impact on reasoning ability}
\label{appendix:reasoning}
To further investigate the impact of \ourmethod{} on more complex reasoning abilities, which was a concern raised in previous reviews, we evaluate the model on the 2WikiMultiHopQA dataset, a benchmark designed to test multi-hop reasoning capabilities through questions requiring connecting information from multiple sources.

The results on 2WikiMultiHopQA (Table \ref{tab:multihop_qa_results}) show a similar trend to MMLU regarding the influence of $\alpha$. The highest EM and F1 scores are achieved when $\alpha$ is close to 0 (e.g., $\alpha\in[0.0,0.7]$). As $|\alpha|$ increases, indicating stronger steering towards either contextual or parametric knowledge, there is a gradual decline in multi-hop reasoning performance. For instance, at $\alpha=+2.0$, the EM score drops to 46.62 from a peak of 54.50. However, it is crucial to note that even at these more extreme $\alpha$ values, the performance of the 72B model (e.g., 46.62 EM at $\alpha=+2.0$) remains significantly higher than that of a much smaller 3B model (26.37 EM), which struggles with the inherent complexity of the task. This suggests that while very strong steering can impact complex reasoning, the \ourmethod{} framework, within a moderate range of $\alpha$, allows for effective context sensitivity adjustment while largely preserving the sophisticated reasoning capabilities of the large model.

\begin{table}[h!]
    \centering
    \resizebox{\linewidth}{!}{ 
        \begin{tabular}{lcc}
            \toprule[1.5pt]
            \textbf{Alpha ($\alpha$)} & \textbf{EM Score} & \textbf{F1 Score} \\
            \midrule[1pt]
            -2.0       & 48.00   & 59.08   \\
            -1.5       & 52.00   & 62.54   \\
            -1.0       & 53.50   & 64.33   \\
            -0.7       & 54.50   & 64.99   \\
            -0.5       & 54.50   & 64.67   \\
            \rowcolor{lightgray}
            \textbf{72B ($\alpha=0$)}  & \textbf{54.50}   & \textbf{64.78}   \\ 
            +0.5       & 53.63   & 64.11   \\
            +0.7       & 53.37   & 63.79   \\
            +1.0       & 52.62   & 62.89   \\
            +1.5       & 50.75   & 60.67   \\
            +2.0       & 46.62   & 56.60   \\
            \midrule[1pt]
            \rowcolor{lightgray}  
            \textbf{3B (baseline)} & \textbf{26.37}   & \textbf{38.35}   \\
            \bottomrule[1.5pt]
        \end{tabular}
    }
    \caption{Performance (EM and F1 scores) of Qwen steered by \ourmethod{} on the 2WikiMultiHopQA multi-hop reasoning benchmark for different $\alpha$ values. Results for a 3B baseline model are also shown for comparison.}
    \vspace{-3mm}
    \label{tab:multihop_qa_results}
\end{table}

The performance decline observed on both MMLU and 2WiKiMultiHopQA suggests that extreme sensitivity adjustments may disrupt the target model's fundamental reasoning patterns, highlighting the importance of maintaining a balanced $\alpha$ range that preserves core competencies while enabling effective knowledge adaptation. Notably, even within this kind-of-broad range, the target 72B model consistently outperforms the 7B/3B proxy models by significant margins (average $+8.67\%$ on MMLU, and substantially higher EM/F1 on 2WikiMultiHopQA), demonstrating that our framework successfully leverages the large model's superior general ability and reasoning capacity while achieving precise context sensitivity control. These findings collectively indicate that strategic $\alpha$ selection can achieve an effective equilibrium between contextual adaptability and model capability preservation, fulfilling our framework's dual objectives of precise knowledge steering and performance maintenance.

\section{\ourmethod{} results on Gemma-2-27b-it}
\label{appendix:transfer}

To further substantiate the transferability of our \ourmethod{} framework and demonstrate its generalization capabilities across diverse LLM architectures, we extended our empirical validation to the Gemma-2 model family. For these experiments, Gemma-2-27b-it was utilized as the target large language model, with its smaller counterpart, Gemma-2-2b-it, serving as the proxy model maintaining the ratio of the size of the target model to the proxy model at approximately 10 to 1. We evaluated performance on both the PopQA and MuSiQue datasets, maintaining the same experimental setup and metrics as used for the Llama-3 and Qwen2.5 models. The comprehensive results for the Gemma-2-it models are presented in Table \ref{tab:transfer-results}.

It is noteworthy that the Gemma-2-27b-it model exhibits a relatively strong baseline performance compared to the Llama-3 and Qwen2.5 models evaluated earlier. Despite this higher baseline, \ourmethod{} consistently delivered substantial and leading improvements across both datasets. The successful application of \ourmethod{} to the Gemma-2 architecture, which differs from the previously tested models, provides compelling evidence for the framework's broad applicability and robust generalization. These results effectively address concerns regarding transferability, highlighting \ourmethod{} as a versatile solution for steering knowledge sensitivity in large language models.

\begin{table*}[t]
\centering
\small
\setlength{\tabcolsep}{3pt}
\renewcommand{\arraystretch}{1.2}
\resizebox{\textwidth}{!}{

\begin{tabular}{lccccccccc}
\toprule
\multirow{2}{*}{\textbf{Methods}} & \multicolumn{2}{c}{\textbf{Degree of Perturbation}(in \%)} & \multicolumn{2}{c}{\textbf{Contextual Detail}(in \%)} & \multicolumn{3}{c}{\textbf{Popularity}(in \%)} & \multirow{2}{*}{\textbf{Sensitivity Score}} \\
\cmidrule(lr){2-3} \cmidrule(lr){4-5} \cmidrule(lr){6-8}
& \textbf{rank 1} & \textbf{rank 2} & \textbf{rank 1} & \textbf{rank 2} & \textbf{rank 1} & \textbf{rank 2} & \textbf{rank 3}\\
\midrule

\multicolumn{9}{l}{\textbf{\uline{\textit{PopQA $\bullet$ Gemma-2-it}}}} \\ [3pt] 
\addlinespace[3pt]
\rowcolor{lightblue_}
Origin & 82.81 & 52.42 & 81.71 & 53.52 & 69.00 & 67.89 & 66.02 & 64.49 &\\
\rowcolor{lightblue_}  
\textsc{Prompt} & 87.44 (\textcolor{magenta}{+4.63}) & 68.28 (\textcolor{magenta}{+15.86}) & \textbf{85.24} (\textcolor{magenta}{+3.53}) & 70.48 (\textcolor{magenta}{+16.96}) & 77.00 (\textcolor{magenta}{+8.00}) & 77.59 (\textcolor{magenta}{+9.70}) & 78.96 (\textcolor{magenta}{+12.94}) & 76.30 (\textcolor{magenta}{+11.81}) & \\
\rowcolor{lightblue_}  
\rowcolor{lightblue_} 
\textsc{CAD} & 87.88 (\textcolor{magenta}{+5.07}) & 66.96 (\textcolor{magenta}{+14.54}) & 88.54 (\textcolor{magenta}{+6.83}) & 66.29 (\textcolor{magenta}{12.77}) & 76.33 (\textcolor{magenta}{+7.33}) & 77.92 (\textcolor{magenta}{+10.03}) & 77.99 (\textcolor{magenta}{+11.97}) & 75.37 (\textcolor{magenta}{+10.88}) & \\
\rowcolor{lightblue_}  
\textsc{COIECD} & 84.14 (\textcolor{magenta}{+1.33}) & 54.62 (\textcolor{magenta}{+2.20}) & 82.37 (\textcolor{magenta}{+0.66}) & 56.38 (\textcolor{magenta}{+2.86}) & 71.00 (\textcolor{magenta}{+2.00}) & 69.56 (\textcolor{magenta}{+1.67}) & 67.63 (\textcolor{magenta}{+1.61}) & 66.38 (\textcolor{magenta}{+1.89}) & \\
\rowcolor{lightblue_}  
\textsc{CSKS} & \textbf{88.98} (\textcolor{magenta}{+6.17}) & \textbf{70.70} (\textcolor{magenta}{+18.28}) & 84.80 (\textcolor{magenta}{+3.09}) & \textbf{74.88} (\textcolor{magenta}{+21.36}) & \textbf{81.33} (\textcolor{magenta}{+12.33}) & \textbf{79.26} (\textcolor{magenta}{+11.37}) & \textbf{78.96} (\textcolor{magenta}{+12.94}) & \textbf{80.47} (\textcolor{magenta}{+15.98}) & \\
\midrule

\multicolumn{9}{l}{\textbf{\uline{\textit{MusiQue $\bullet$ Gemma-2-it}}}} \\ [3pt] 
\addlinespace[3pt]
\rowcolor{lightgrey}  
Origin & 85.13 & 40.42 & 72.76 & 52.86 & 68.16 & 60.71 & 59.62 & 59.01 &\\
\rowcolor{lightgrey}  
\textsc{Prompt} & 88.95 (\textcolor{magenta}{+3.82}) & 50.00 (\textcolor{magenta}{+9.58}) & 75.10 (\textcolor{magenta}{+2.34}) & 63.90 (\textcolor{magenta}{+11.04}) & 73.63 (\textcolor{magenta}{+5.47}) & 67.20 (\textcolor{magenta}{+6.49}) & 67.70 (\textcolor{magenta}{+8.08}) & 66.60 (\textcolor{magenta}{+7.59}) & \\
\rowcolor{lightgrey}  
\rowcolor{lightgrey}  
\textsc{CAD} & 88.32 (\textcolor{magenta}{+3.19}) & 53.61 (\textcolor{magenta}{+13.19}) & 78.51 (\textcolor{magenta}{+5.75}) & 63.48 (\textcolor{magenta}{+10.62}) & 77.17 (\textcolor{magenta}{+9.01}) & 67.85 (\textcolor{magenta}{+7.14}) & 68.01 (\textcolor{magenta}{+8.39}) & 67.89 (\textcolor{magenta}{+8.88}) & \\
\rowcolor{lightgrey}  
\textsc{COIECD} & 85.56 (\textcolor{magenta}{+0.43}) & 42.55 (\textcolor{magenta}{+2.13}) & 73.82 (\textcolor{magenta}{+1.06}) & 54.35 (\textcolor{magenta}{+1.49}) & 69.45 (\textcolor{magenta}{+1.29}) & 61.36 (\textcolor{magenta}{+0.65}) & 61.49 (\textcolor{magenta}{+1.87}) & 60.43 (\textcolor{magenta}{+1.42}) & \\
\rowcolor{lightgrey}  
\textsc{CSKS} & \textbf{93.63} (\textcolor{magenta}{+8.50}) & \textbf{70.85} (\textcolor{magenta}{+30.43}) & \textbf{83.82} (\textcolor{magenta}{+11.06}) & \textbf{80.68} (\textcolor{magenta}{+27.82}) & \textbf{84.89} (\textcolor{magenta}{+16.73}) & \textbf{80.19} (\textcolor{magenta}{+19.48}) & \textbf{81.67} (\textcolor{magenta}{+22.05}) & \textbf{80.75} (\textcolor{magenta}{+21.74}) & \\
\midrule
\end{tabular}
}
\caption{Performance of \ourmethod{} and baseline methods on PopQA and MuSiQue datasets using Gemma-2-27b-it as the target LLM and Gemma-2-2b-it as the proxy model. Results show accuracy (in \%) for different ranks of perturbation, contextual detail, and popularity, along with the overall Sensitivity Score. Improvements by \ourmethod{} over the Origin are shown in magenta.}
\label{tab:transfer-results}
\end{table*}

\section{Prompts used to generate our synthesized dataset}
\label{appendix:prompts}
\begin{figure}[t]
\centering
\begin{minipage}{0.5\textwidth} 
    \centering
    \includegraphics[width=\textwidth]{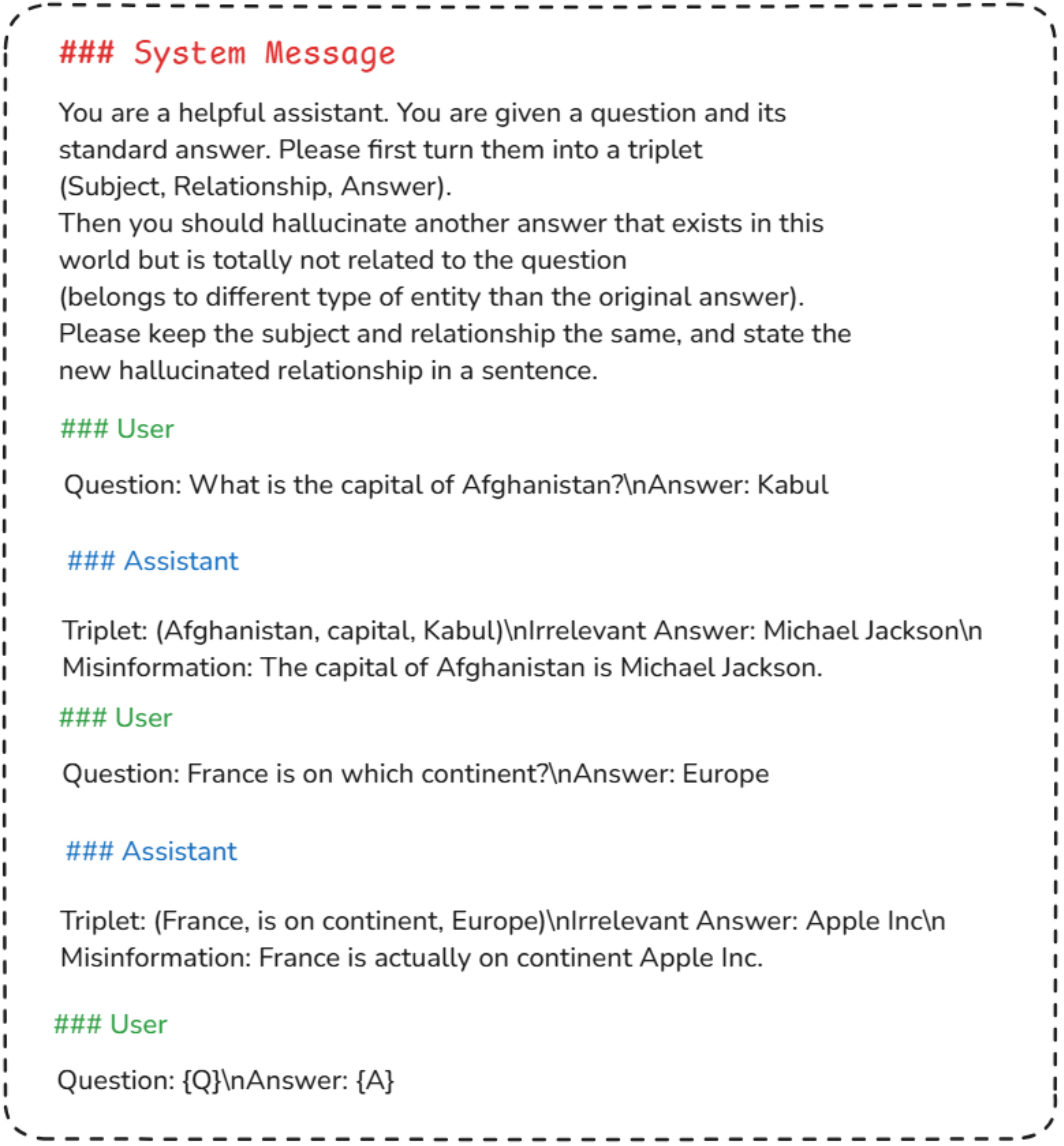}  
    \caption{The prompt we use to ask gpt to make a slight permutation.}
    \label{fig:f7}
\end{minipage}
\end{figure}

\begin{figure}[t]
\centering
\begin{minipage}{0.5\textwidth} 
    \centering
    \includegraphics[width=\textwidth]{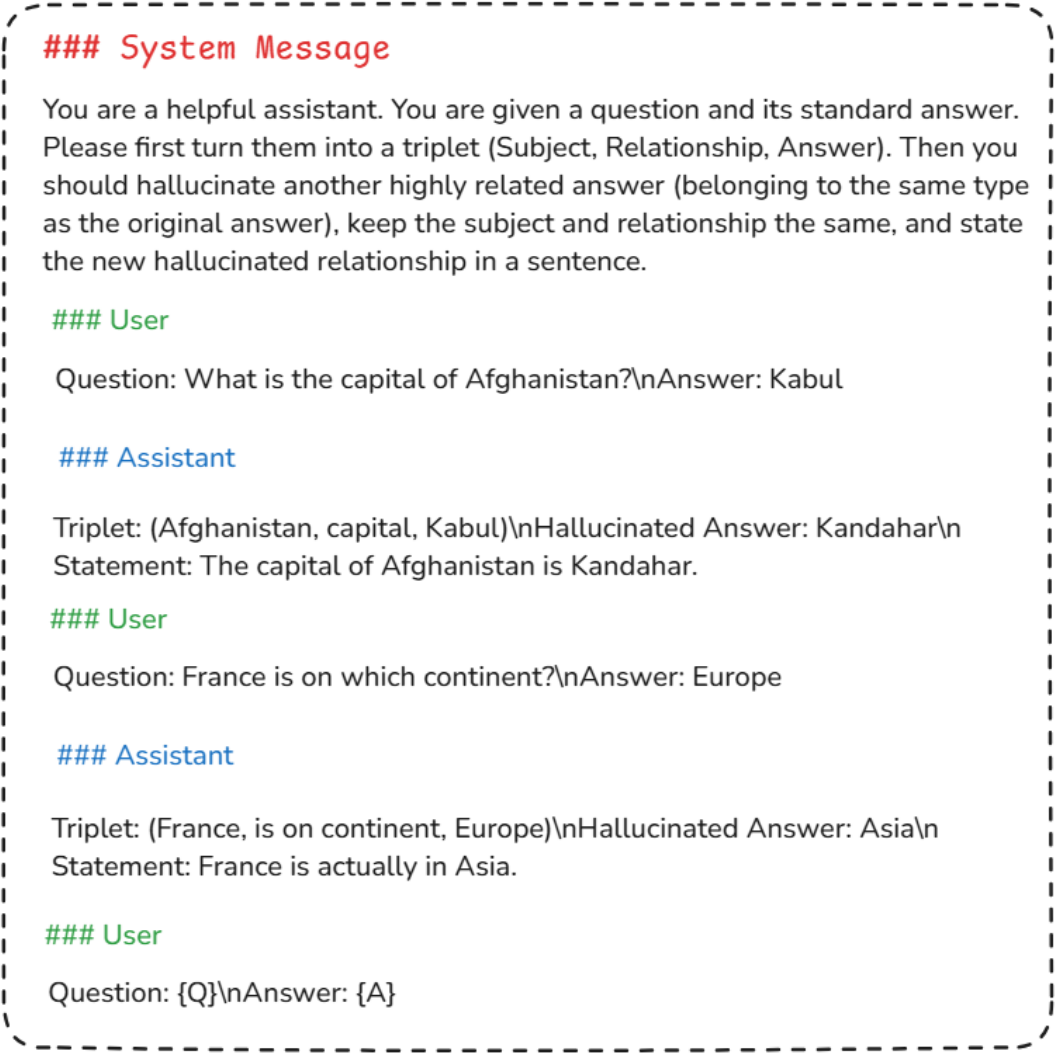}  
    \caption{The prompt we use to ask gpt to make a siginificant permutation.}
    \label{fig:f8}
\end{minipage}
\end{figure}

\begin{figure}[h]
\centering
\begin{minipage}{0.5\textwidth} 
    \centering
    \includegraphics[width=\textwidth]{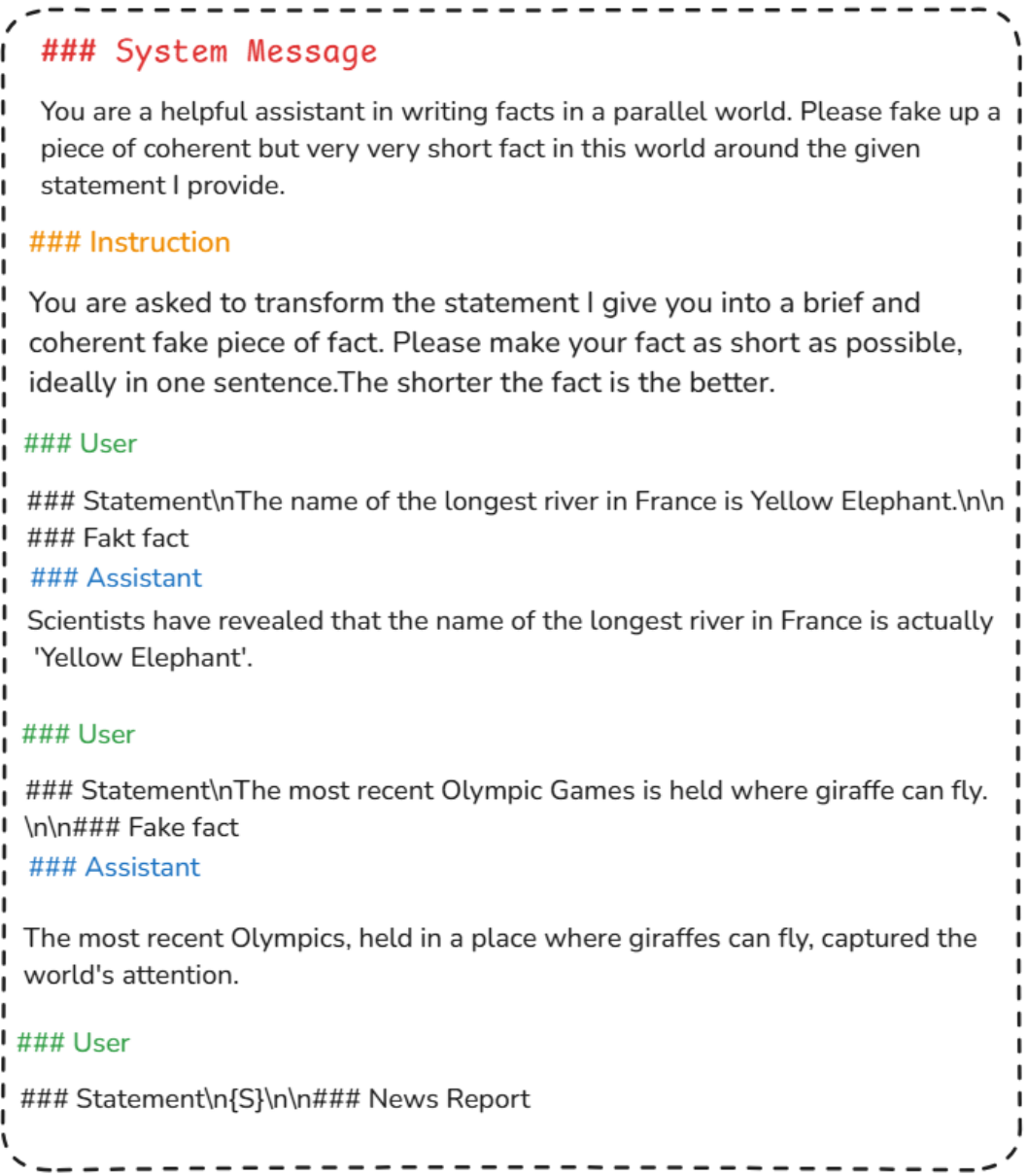}  
    \caption{The prompt we use to ask gpt to generate a short context.}
    \label{fig:f9}
\end{minipage}
\end{figure}

\begin{figure}[t]
\centering
\begin{minipage}{0.5\textwidth} 
    \centering
    \includegraphics[width=\textwidth]{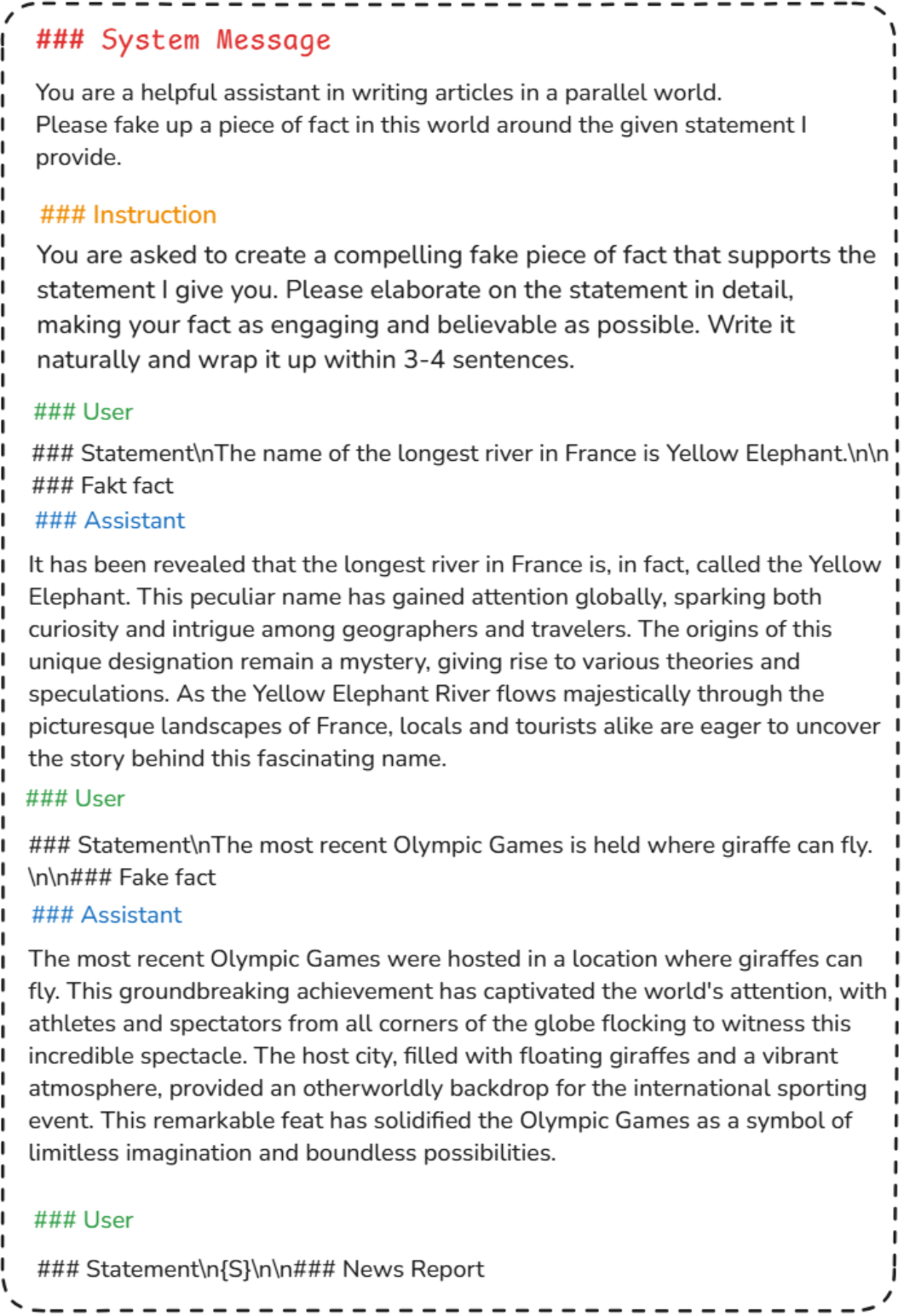}  
    \caption{The prompt we use to ask gpt to generate a long context.}
    \label{fig:f10}
\end{minipage}
\end{figure}

Figure \ref{fig:f7} - Figure \ref{fig:f10} show the prompts used to generate the features for different dimensions of our dataset.

\end{document}